\pdfoutput=1
\documentclass[11pt]{article}

\usepackage[preprint]{acl}

\usepackage{times}
\usepackage{latexsym}
\usepackage{tabularx}
\usepackage{booktabs}
\usepackage{makecell}

\usepackage[T1]{fontenc}
\usepackage[utf8]{inputenc}

\usepackage{microtype}

\usepackage{inconsolata}

\usepackage{graphicx}
\usepackage{amsmath}

\title{Emotional RAG LLMs: \\Reading Comprehension for the Open Internet}

\author{Benjamin Reichman\textsuperscript{1}, Adar Avsian\textsuperscript{1},   Kartik Talamadupula\textsuperscript{2}$^{\ast}$, \\   \textbf{Toshish Jawale\textsuperscript{3}$^{\ast}$, Larry Heck\textsuperscript{1}} \\
\textsuperscript{1}AI Virtual Assistant (AVA) Lab,
Georgia Institute of Technology \\
\textsuperscript{2}Wand AI,
\textsuperscript{3}Invoca \\
  \texttt{\{bzr,aavsian3,larryheck\}@gatech.edu} \\
  \texttt{kartik@wand.ai}, \texttt{toshish.jawale@gmail.com}
}

\begin{document}
\maketitle
\footnotetext[1]{Work done while at Symbl.AI}
\begin{abstract}
Queries to large language models (LLMs) can be divided into two parts: the instruction/question and the accompanying context. The context for retrieval-augmented generation (RAG) systems in most benchmarks comes from Wikipedia-like texts written in a neutral and factual tone. However, real-world RAG applications often retrieve internet-based text with diverse tones and linguistic styles, posing challenges for downstream tasks. This paper introduces (a) a dataset that transforms RAG-retrieved passages into emotionally inflected and sarcastic text, (b) an emotion translation model for adapting text to different tones, and (c) a prompt-based method to improve LLMs' pragmatic interpretation of retrieved text.
\end{abstract}

\section{Introduction}
Over the past few years, large language models (LLMs) have vastly expanded in their scope of use. However, LLMs have shortcomings, such as a tendency to generate hallucinated content—confidently providing incorrect or fabricated information \cite{chatgpt_hallucinations}. Since LLMs have finite number of parameters and knowledge is inherently a long-tail problem, it is infeasible for LLMs to memorize all the information necessary to answer every possible query \cite{longtail_knowledge}. Additionally, models have a ``knowledge cutoff'' date, which refers to knowledge that occurs after their pretraining starts. Together, these factors highlight the need for augmenting LLMs with external knowledge sources.

Retrieval-augmented generation (RAG) mitigates LLM hallucinations by retrieving external knowledge \cite{rag}. While benchmarks often use neutral, fact-based Wikipedia text, real-world deployments pull from the internet, where tone and intent vary. Retrieved passages may contain sarcasm, irony, or emotional inflections, requiring pragmatic reasoning to interpret meaning correctly. Without this ability, LLMs risk misreading context, leading to incorrect or harmful outputs \cite{googleaioverview,googleaioverview2}.

Human communication relies on both literal and non-literal meaning, requiring pragmatic reasoning to interpret emotionally inflected, sarcastic, or ironic text. Pragmatics studies how context shapes meaning, a critical skill for LLMs processing retrieved internet text, which varies in tone and intent.

Prior work has evaluated LLMs' pragmatic reasoning in multiple-choice tasks and conversational settings \cite{pope}, but these focus on interpreting queries. In contrast, RAG introduces a query-context split, where an LLM must read pragmatically complex retrieved passages. This work examines how well LLMs handle pragmatic reading in retrieved contexts, particularly when emotional tone and sarcasm distort meaning.

To address variability in emotions and linguistic tropes in reading comprehension for the RAG task, the Reading with Intent task is introduced. This work makes the following contributions:
\begin{enumerate}
    \item Create a dataset of retrieved contexts synthetically transformed into $11$ distinct emotions.
    \item Develop an emotion translation model capable of adapting text to specified emotional tones.
    \item Develop a prompt-based approach to pragmatically reading provided context.
\end{enumerate}

\section{Related Works}
\textbf{Sentiment Analysis:} Numerous methods have been proposed to identify emotions in text \cite{prabowo2009sentiment,medhat2014sentiment,wadawadagi2020sentiment,wankhade2022survey}, but research on reading comprehension involving diverse emotional tones remains limited.

\textbf{Style Transfer:} Previous work on emotion translation often uses style-transfer techniques. One method removes translating text into another language to remove the text's style and translates back using a style-specific decoder \cite{prabhumoye-etal-2018-style}. Another employs zero-shot learning, prompting LLMs with multiple style-transfer examples before asking them to generate novel transformations \cite{reif-etal-2022-recipe}. Several studies have explored style-based emotion transfer \cite{li-etal-2019-domain,qi-etal-2021-mind,Shen2017StyleTF,Yang2018UnsupervisedTS,mir-etal-2019-evaluating}, but they typically focus on binary or coarse-grained emotions (e.g., positive vs. negative). In contrast, this work takes a data-centric approach, synthetically generating a bi-text corpus spanning 11 distinct emotions, allowing for direct emotion translation rather than a style transfer approach.

\textbf{Sarcasm Detection:} Sarcasm detection methods use techniques like CNN-based feature extraction \cite{Poria2016ADL}, graph learning \cite{Lou2021AffectiveDG}, and commonsense reasoning \cite{commonsensesarcasm}, trained on datasets such as SARC and iSarcasm \cite{sarc,isarcasm}. Sarcasm detection datasets, while useful for classification tasks, are not usable for the pragmatic reading task this paper addresses. The datasets are comprised of one-off posts about no topic in particular. The dataset needed and created in this paper are topically-relevant sarcastic passages that can be used to answer a query. While prior methods for detection are effective, these approaches do not integrate sarcasm understanding into reading comprehension tasks. Our work builds on these foundations by addressing sarcasm’s impact on retrieved passage interpretability.

\textbf{Sarcasm Generation: } There have been a few prior works on sarcasm generation. One approach employs logical representations to transform sentences into sarcastic versions \cite{oprea-etal-2021-chandler}. Another method reverses the sentiment polarity (valence) of the input sentence and uses the commonsense reasoning framework COMET to generate sarcastic context that aligns with the transformed statement \cite{chakrabarty-etal-2020-r}.

\textbf{Pragmatic Reading:} Pragmatics studies how non-literal meaning arises from context. Many works have explored implicature classification using language models \cite{implicature1,implicature2,implicature3}, while others have studied LLMs selecting the correct pragmatic response in multiple-choice tasks involving sarcasm, metaphor, and violations of Gricean maxims \cite{mulchoicepragmatics1}. Recently there has been work on generating pragmatically appropriate responses in conversational turns \cite{pope}. 

To the best of our knowledge, this is the first work applying pragmatics to reading retrieved-context in a RAG setting. We focus specifically on pragmatic reading of sarcasm, a crucial ability for LLMs given the prevalence of non-literal language on the internet. Without this capability, models risk misinterpreting sarcasm and generating harmful or misleading outputs \cite{googleaioverview,googleaioverview2}.

\section{Dataset Creation}
\label{sec:dataset}

\begin{figure*}[htb!]
\centering
\includegraphics[width=0.85\textwidth]{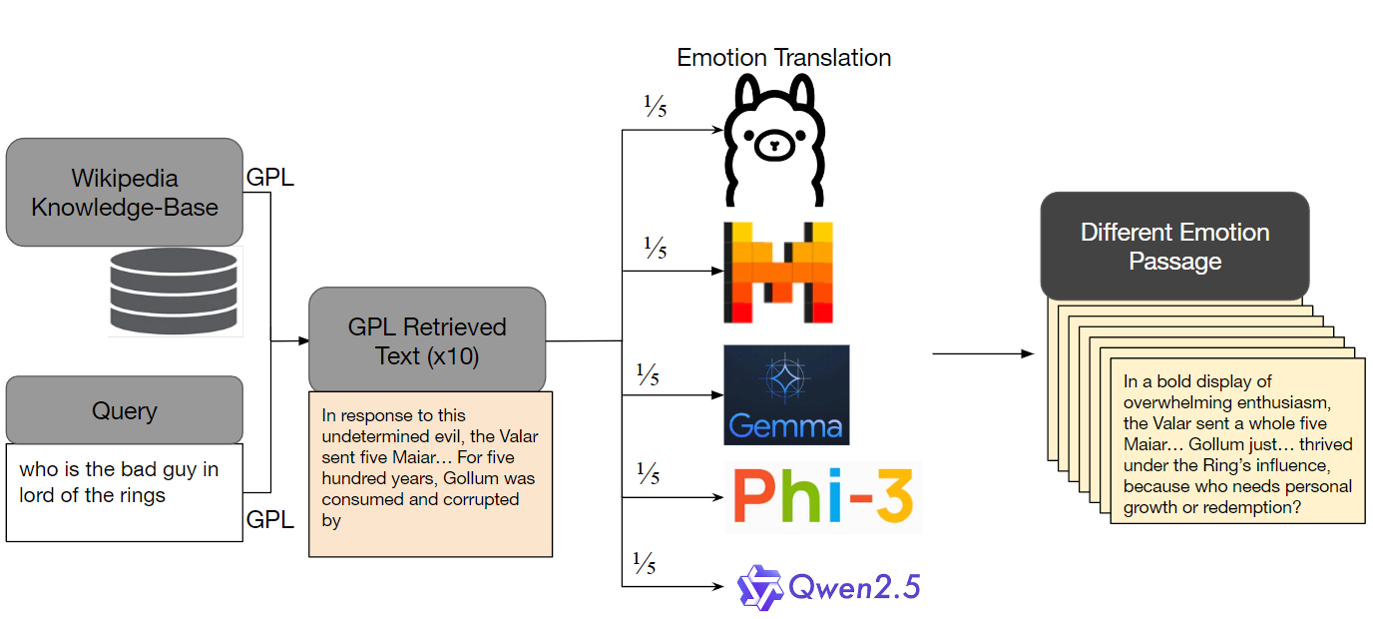}
\caption{Synthetic data generation process.}
\label{fig:dataset_creation}
\end{figure*}

To study the problem of reading and answering questions over emotionally inflected text, an open-domain question-answering (QA) dataset with a retrieval corpus specifically tailored to this task is necessary. While various datasets for sentiment analysis exist, they are typically limited to classification tasks and do not address the complexities of emotion understanding in the context of QA. This work, in contrast, focuses on how emotion, particularly sarcasm, affects the reading and answering of questions. 
% To explore this problem a dataset needed to be generated.
Building a dataset for this intention-aware reading task requires several steps:

\begin{enumerate}
    \item Identifying a set of open-domain questions.
    \item Retrieving passages for those questions.
    \item Adding emotional inflection and sarcasm to the retrieved passages.
    \item Integrating synthetic passages into the dataset.
\end{enumerate}

% \begin{itemize}
    % \item Overview
    % \begin{itemize}
    %     \item To study the problem of reading emotionally inflected text, a dataset of emotionally inflected text was needed.
    %     \item Various datasets for sentiment analysis exist.
    %     \item However, these datasets are usually limited to classification tasks.
    %     \item This paper, however, investigates the use of emotion-understanding while reading for QA.
    %     \item To be able to explore this problem we had to generate the dataset.
    % \end{itemize}
    % \item Steps
    % \begin{itemize}
    %     \item Building a dataset for the intention-aware reading task required several steps.
    %     \begin{enumerate}
    %         \item Identifying a set of open-domain questions for the model to answer
    %         \item Retrieving passages for those questions
    %         \item Adding sarcasm to the retrieved passages.
    %         \item Integrating sarcasm-poisoned passages into the dataset
    %     \end{enumerate}
    % \end{itemize}
% \end{itemize}
\subsection{Open-Domain Questions}

The objective of the dataset is to have a set of questions such that an LLM would be required to read outside passages to provide accurate answers. This requires questions that an LLM would have a low likelihood of being able to answer relying solely on its internal knowledge learned during pretraining, thus necessitating the use of outside knowledge and reading comprehension abilities. We thus chose an open-domain question-answering dataset \cite{nq} that provides a retrieval corpus as well as ground truth retrieved passages as the base for our new dataset.

% \begin{itemize}
%     \item The objective of the dataset is to have an LLM answer questions that require it to read passages. 
%     \item This requires questions that LLM would have a low likelihood of being able to answer relying solely on its internal knowledge, thus requiring outside knowledge and therefore reading comprehension abilities.
%     \item We thus chose an open-domain dataset that provides a retrieval corpus and ground truth retrieved passages as a base for our dataset. 
% \end{itemize}
\subsection{Retrieval}
\label{sec:retrieval}

Although retrieval is a critical portion of the RAG pipeline, it is not the focus of this paper. Therefore, an off-the-shelf SOTA dense retrieval method, specifically GPL \cite{gpl} was used. Each query in the NQ dataset was embedded with the query encoder. Each passage in the NQ Wikipedia retrieval corpus was embedded with the passage encoder. For each query in the dataset the top-$200$ passages that maximized the inner product score were retrieved as illustrated in Figure~\ref{fig:dataset_creation}. These retrieved passages formed the base from which our dataset was derived.

% \begin{itemize}
%     \item Although retrieval is a critical portion of the RAG pipeline, it is not the object of our study here.
%     \item We therefore used an off-the-shelf state-of-the-art dense retrieval method. 
%     \item With GPL we retrieved the top-200 passages for each question in NQ's validation set.
%     \item The retrieved passages will form a base from which our dataset is derived.
% \end{itemize}
\subsection{Emotion Distortion}

With the passages retrieved, the next step was generating emotionally altered versions. To ensure scalability and consistency, we synthetically transformed passages into different emotions instead of manually rewriting them.

Figure \ref{fig:dataset_creation} illustrates how each passage was converted into 11 distinct emotional or linguistic variations: anger, condescension, disgust, envy, excitement, fear, happiness, humor, sadness, sarcasm, and surprise. To ensure diversity, each passage was randomly assigned to one of five LLMs: Llama 3, Qwen 2.5, Phi-3, Gemma, and Mistral-7B. The variety of models captures different linguistic nuances and stylistic variations, leading to a richer and more representative instantiation of each emotion. Appendix \ref{sec:prompts} contains each of the prompts used to create the synthetic dataset.

For sarcasm, we generated two types of passages: (1) Sarcastic factually correct passages and (2) Sarcastic factually distorted passages. This distinction allows us to examine how LLMs use pragmatics to interpret retrieved information, particularly when tone and factuality conflict. By analyzing how models process these cases, we gain insight into their ability to navigate pragmatic cues when assessing meaning and reliability.

For type (1), we prompted an LLM to rewrite the retrieved passage with sarcasm while preserving factual correctness. For type (2), we first prompted an LLM to distort general factual details within the passage, and if the passage contained the ground truth, the prompt explicitly instructed the model to alter that specific fact as well. The resulting fact-distorted passages were then reprocessed with a sarcasm-inducing prompt. This two-step approach ensures our dataset captures both accurate and misleading sarcastic text, enabling evaluations of how LLMs handle different types of sarcastic content.

\subsection{Integration}
\label{sec:integration}

% \begin{figure*}[htb!]
% \centering
% \includegraphics[width=\textwidth]{paper8_figure1.PNG}
% \caption{Synthetic data generation process.}
% \label{fig:dataset_creation}
% \end{figure*}
After generating the synthetic passages, the next step was integrating them into the existing retrieval results. This work focuses on integrating the sarcastic passages. Future work will integrate the other emotions into the neutral dataset. To evaluate reading comprehension methods under different conditions, three test datasets were created.

The first dataset replaces all original passages with their factually-accurate sarcastic equivalents. This isolates the effect of sarcasm, testing the model’s ability to interpret and respond to altered connotations without factual distortion.

Since real-world retrieval is unlikely to yield only sarcastic passages, the second dataset introduced is only partially sarcastic. Here, 20\% of incorrect passages are replaced with sarcastic versions, and the first two correct passages are paired with a fact-distorted sarcastic passage. This dataset has two variants: one where the sarcastic fact-distorted passage appears before (pre-fix) and one where it appears after (post-fix) the correct passage. These variants test whether passage order affects the model’s ability to discriminate factual reliability.

The third dataset embeds synthetically generated passages into the vector database using the same passage encoder as the retrieval method \cite{gpl}. Queries are then reprocessed using the retrieval pipeline described in Section \ref{sec:retrieval}. This dataset reflects a realistic retrieval distribution, where the prevalence of sarcastic fact-distorted passages is determined by retrieval rankings rather than manual selection.

\section{Dataset Validation}
In this section, we validate the generated dataset through human evaluation. The evaluation assesses whether the transformed text accurately reflects the target emotion and how naturally and convincingly the emotion is conveyed. This ensures that the synthetic dataset aligns with human expectations of emotional expression.

The human evaluation was conducted using Amazon Mechanical Turk (AMT). For each generated emotion, $150$ passages were randomly selected, evenly distributed among the five models used to create the synthetic data. Turkers received these passages as part of a HIT, with each task containing six passages: two example passages demonstrating the target emotion, three newly generated passages from the target emotion, and one passages from different emotions. The example passages served as references to help annotators recognize the intended emotion.

Each HIT was reviewed by three Turkers, who were asked to evaluate the four passages, assigning two ratings per passage using a five-point Likert scale. The first rating assessed whether the passage conveyed the target emotion, while the second measured how realistically the emotion was expressed. The inclusion of a passage from a different emotions served as a sanity check to ensure that Turkers were not selecting responses randomly. Turkers who failed to distinguish between emotions in the sanity check passages were filtered out to maintain annotation quality.

\begin{figure}[ht!]
\centering
\includegraphics[width=\columnwidth]{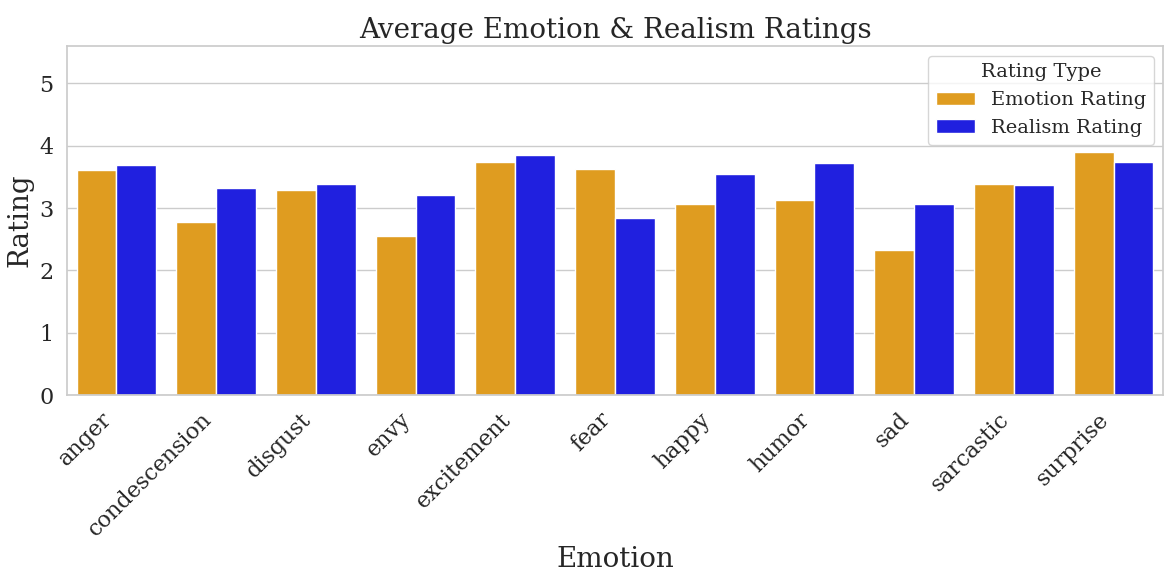}
\caption{Human evaluation of how the synthetic emotions are written and their realism.}
\label{fig:realism}
\end{figure}

Figure \ref{fig:realism} presents the average ratings assigned by Turkers for each passage in the synthetically generated dataset. On average, the crowdworkers found that the emotions were written in a way that clearly conveyed the emotion (mean: $3.22/5$) in a realistic manner (mean: $3.44/5$). Appendix \ref{app:datasetanalysis} provides a per-model breakdown, identifying which models produce the most and least realistic emotional transformations. Additionally, the appendix includes descriptive statistics of the generated dataset, offering further insight into its linguistic characteristics.

\section{Reading with Intent}
\label{sec:readintent}

The goal of our work is to enable LLMs to recognize the emotional and pragmatic intent of passages when answering open-domain questions. To explore this, we created a dataset with sarcasm-infused passages (Section~\ref{sec:dataset}). This section presents our framework, Reading with Intent, which improves an LLM’s ability to interpret connotation in retrieved text.

We investigate three approaches: (1) Prompt-Based Pragmatic Reading (Section~\ref{sec:readingprompt}), where we explicitly guide the model to process connotation; and (2) Intent Tags (Section~\ref{sec:intenttags}), which further enhances the prompt-based approach with learned intent annotations, (3) Intent Neutralization (Section~\ref{sec:intentneutralization}), where we strip pragmatics from text.

% \begin{itemize}
%     \item Overview
%     \begin{itemize}
%         \item The goal of our work is for LLMs to be cognizant of the emotional-intent of the passages it reads when trying to answering open-domain questions. 
%         \item To do this we created a dataset with sarcasm-poisoned passages on which to experiment on.
%         \item In this section, we present our framework, Reading with Intent, which helps models to better understand the connotation of the passages that they are reading.
%         \item In section \ref{sec:readingprompt}, we present a prompt-based approach that improves reading accuracy on the datasets presented in section \ref{sec:dataset}.
%         \item In section \ref{sec:intenttags}, we further enhance our prompt-based approach with trained intent tags.
%         \item (Finally, in section \ref{sec:instructiontuning}, we present a model tuned to better read and interpret retrievals that include sarcastic passages.)
%     \end{itemize}
% \end{itemize}
\subsection{Prompt-Based Approach}
\label{sec:readingprompt}

The first approach focuses on pragmatics-aware reading guidance for the model.

Previous question-answering approaches prompt the model to read retrieved passages for one of a few reasons: generating a direct answer, producing candidate answers, or summarizing the passage to aid a downstream model in answering. However, these approaches generally overlook an important aspect—prompting the model to pay attention to the connotation of the text.

Without explicit guidance to focus on connotation, the model pays it less attention. However, by instructing the model to consider the intent behind the text, we can effectively refocus its attention on pragmatic meaning. Our proposed system integrates this by explicitly prompting the LLM to analyze connotation. The full prompt used is provided in appendix \ref{app:prompt}.

\subsection{Intent Tags}
\label{sec:intenttags}
Reading with intent has two steps: detecting connotative intent and interpreting text accordingly. Our system separates these tasks by using a fine-tuned sarcasm detector to generate intent tags, which are embedded in the prompt to steer the LLM’s interpretation. Future work could generalize this to broader emotive labels for deeper nuance.

\subsection{Intent Neutralization}
\label{sec:intentneutralization}
The last approach tested was one that bypasses the need for pragmatic reasoning by neutralizing the emotional intent of a passage. To achieve this, we trained an intent translator, a model designed to strip passages of their emotional inflection and present them in a neutral tone.

\begin{table}[t]
{
\small
\centering
\begin{tabularx}{\columnwidth}{ccc}
\toprule
\textbf{} & \makecell{\textbf{Llama 3.1} \\ \textbf{Results}} & \makecell{\textbf{Emotional} \\ \textbf{Translator Results}} \\
\midrule
Average BLEU Score & $1.25$ & $5.82$ \\
% Multiplier & $1$x & $4.87$x \\
Average BLEURT Score & $0.51$ & $0.54$ \\
\bottomrule
\end{tabularx}
}
\caption{Average BLEU and BLEURT scores.}
\label{tab:bleu}
\end{table}

Using the parallel bitext corpus of emotions and implicatures, an intent-translator can be trained to convert text from one emotional tone to another with high fidelity. The objective of the intent-translator is to accurately and fluently adapt the emotional content of a passage while preserving its semantic meaning.

To train the intent-translator, each training example $x$ was prefixed with a prompt $p$ specifying the source emotion and the target emotion. This prompt guides the model in performing the desired transformation. The pretrained language model was fine-tuned to predict the next token in the target emotional tone using a cross-entropy loss:

$\mathcal{L}_{\text{CE}}(\theta) = -\sum_{j=1}^{N} \log p(y_j | <y_{<j}, x; \theta)$

\begin{figure}[ht!]
\centering
\includegraphics[width=\columnwidth]{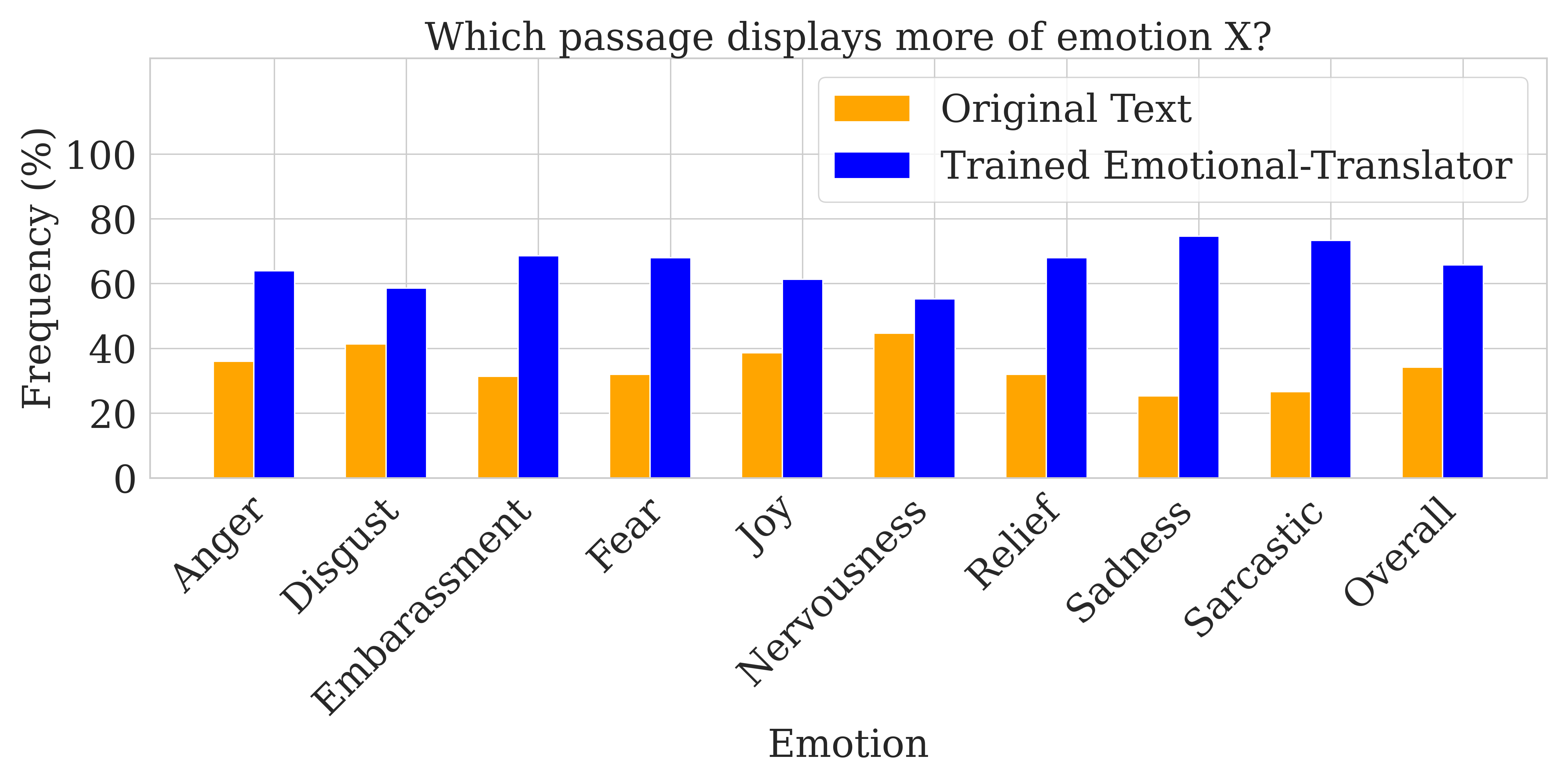}
\caption{Human evaluation of the emotional reconstruction of the human-written text.}
\label{fig:more_emotion_original}
\end{figure}

\begin{figure}[h!]
\centering
\includegraphics[width=0.9\columnwidth]{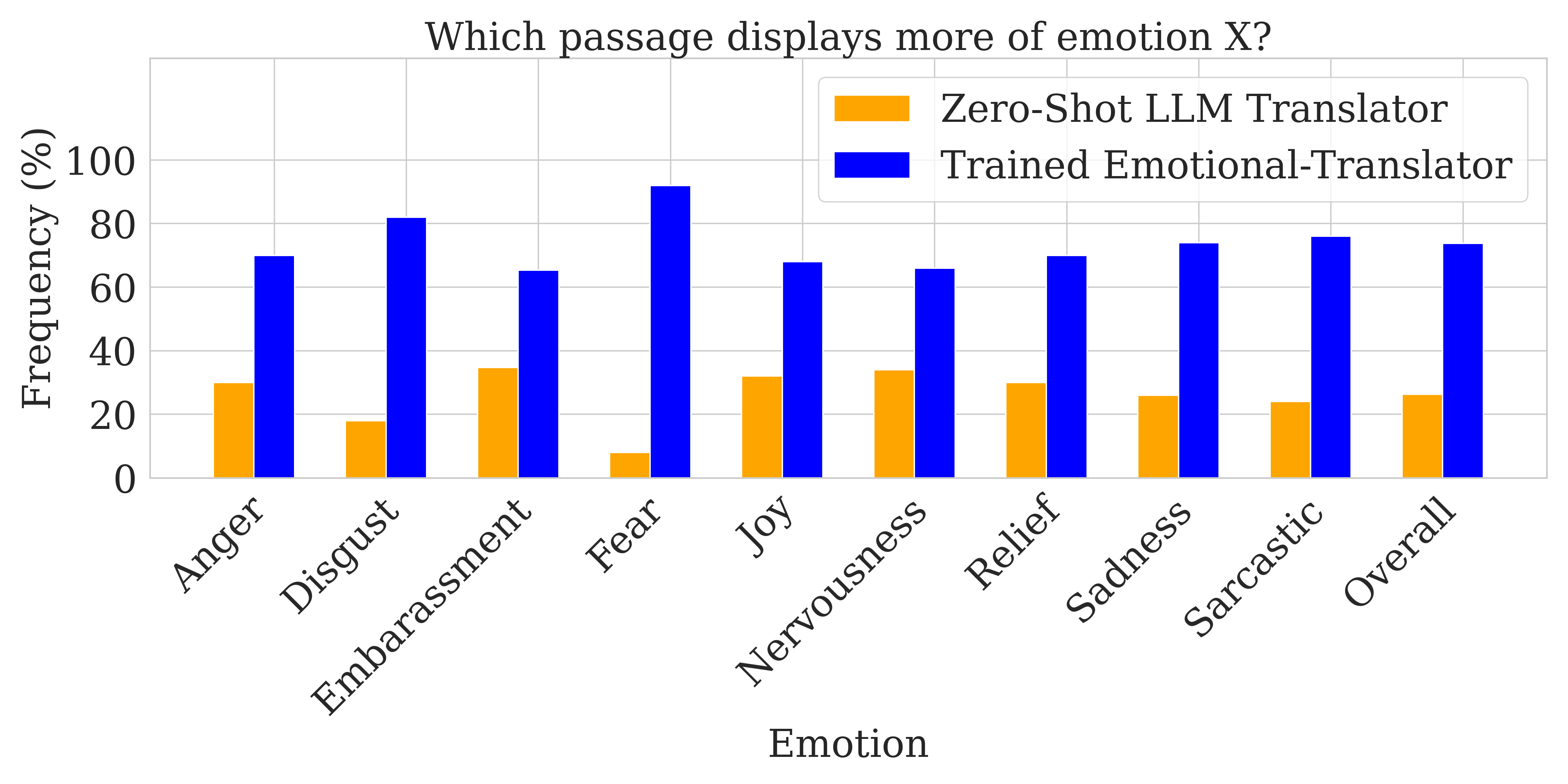}
\caption{Human evaluation of the emotional reconstruction of the human-written text.}
\label{fig:more_emotion_adapter}
\end{figure}

\section{Intent-Translator Evaluation}
\label{app:humevalet}
This section evaluates the intent-translator using both automatic and human metrics across a single task with three criteria. The model was given \textbf{human-written text} labeled by emotion and asked to translate it into a specified target emotion, then back-translate it to the original. Using \textbf{human-authored input} is essential, as it represents both an out-of-distribution challenge—since the model is not fine-tuned on such inputs—and the ultimate use case for intent translation.

Back-translation is necessary as human-written sentiment datasets typically lack paired bitext for emotions, making direct evaluation of emotion translation infeasible. Thus, the translator's performance is assessed based on its ability to complete a round-trip translation, preserving semantic and emotional fidelity across the transformations.

Human evaluations were carried out using Amazon Mechanical Turk. Each sample was viewed by three US-based turkers, and results reflect a majority vote. Humans performed pairwise comparisons between the round-trip translations of both the fine-tuned and unfine-tuned LLMs to assess their ability to reconstruct the factual and emotional content of the original text. They also compared the original input against the fine-tuned intent-translator’s output to judge which better conveyed the intended emotion. Finally, raters directly evaluated the intent-translator’s output—without back-translation—by scoring how well it expressed the target emotion and how realistic the expression was. Inter-rater agreement scores are included in the appendix.

\begin{figure}[h!]
\centering
\includegraphics[width=0.9\columnwidth]{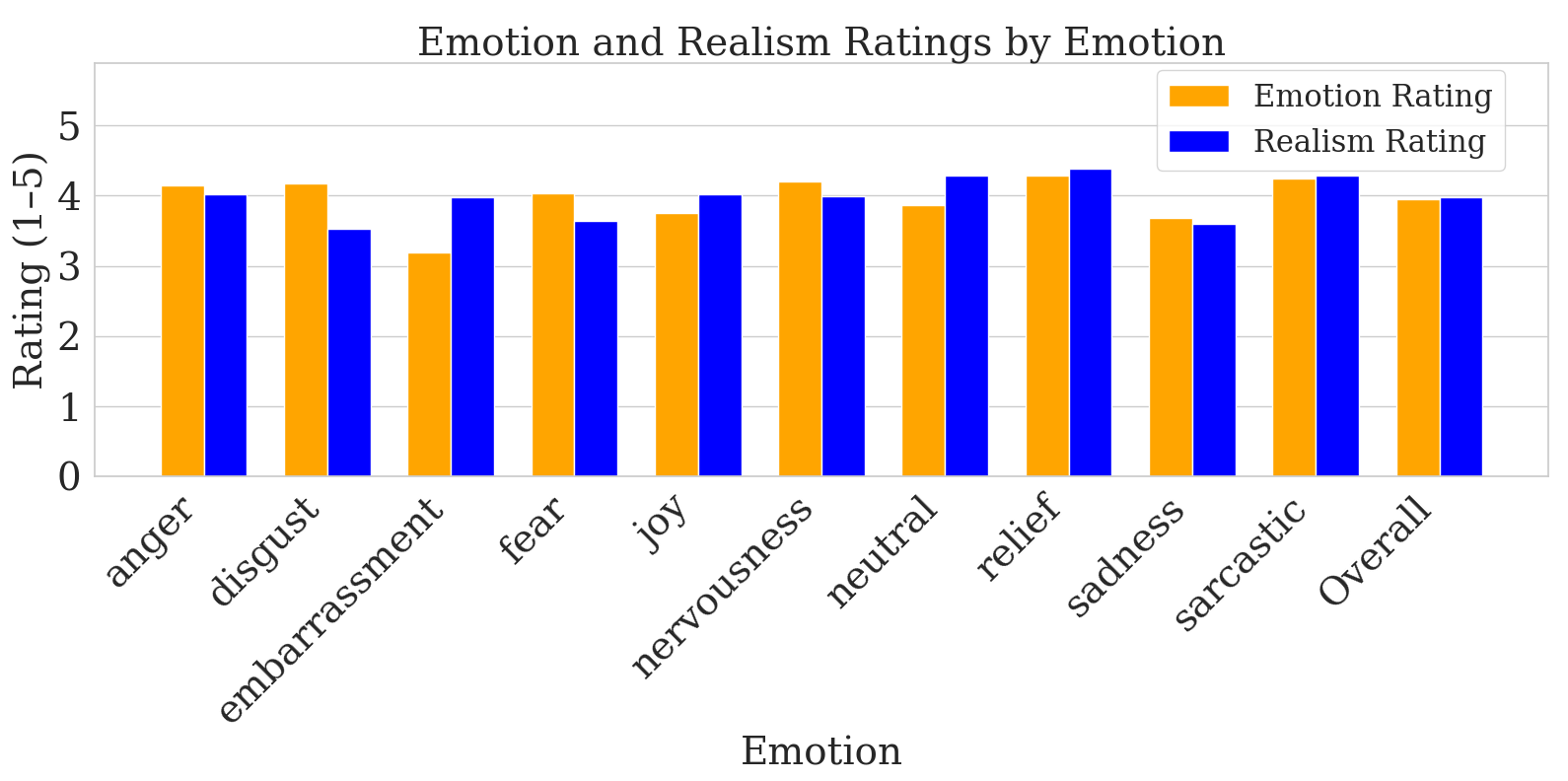}
\caption{Human evaluation of the one-way translation of human-written text.}
\label{fig:emotion_translation_direct}
\end{figure}

\begin{figure}[h!]
\centering
\includegraphics[width=0.9\columnwidth]{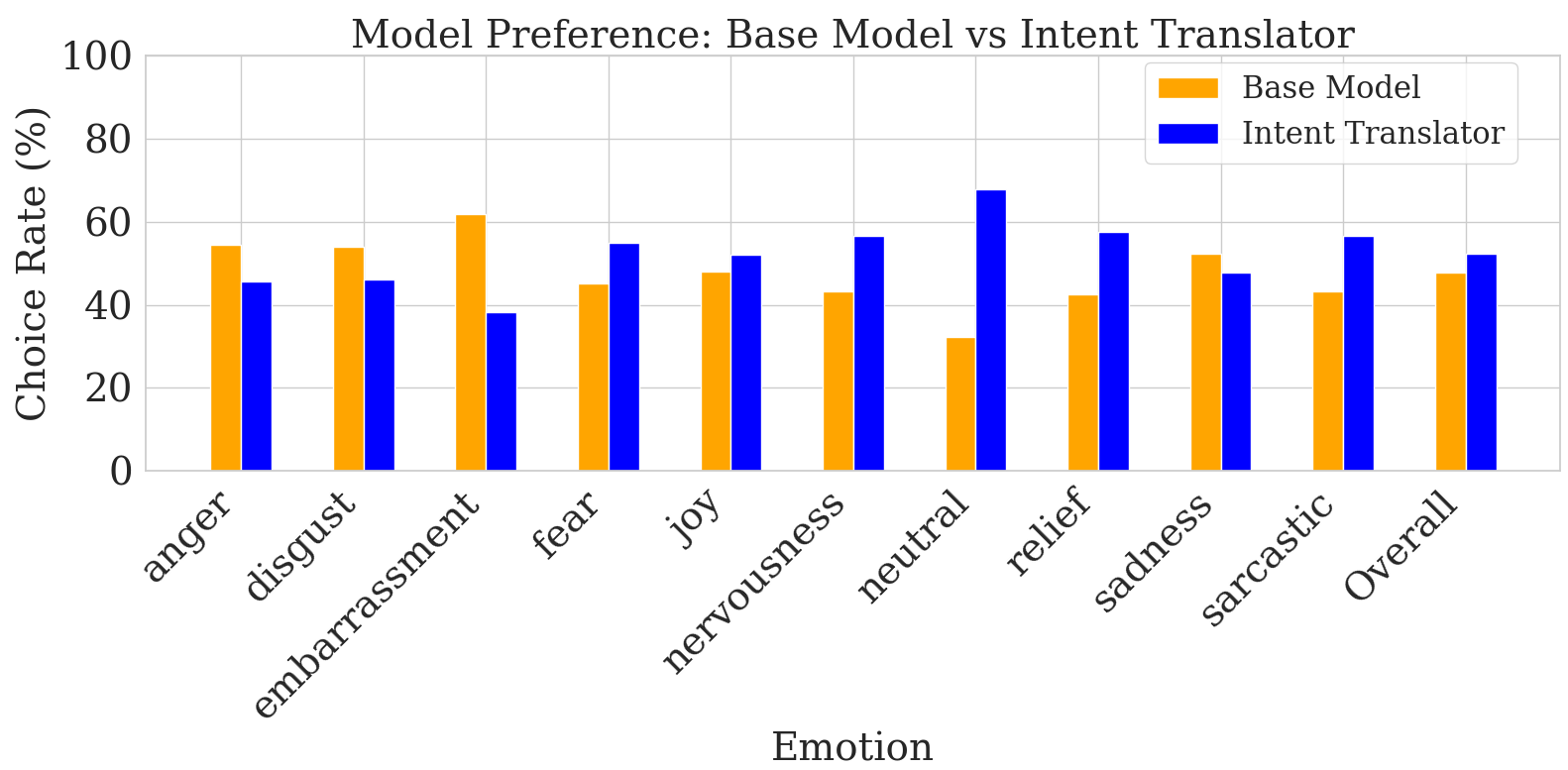}
\caption{Human preference between base model and intent-translator.}
\label{fig:model_preference}
\end{figure}

Two datasets were used as sources of human-written emotional text for the translation experiments: the Go Emotions dataset and the SARC dataset \cite{goemotions,sarc}. The Go Emotions dataset provides fine-grained classification of $28$ emotions with $211,225$ samples sourced from Reddit. The SARC dataset is a dataset dedicated to sarcasm. This dataset has $32,333$ samples of sarcastic text from Reddit. Both datasets are \textbf{human-written}.

From the Go Emotions dataset, eight emotions were sampled: anger, disgust, embarrassment, fear, joy, nervousness, relief, and sadness. Three of these emotions—embarrassment, nervousness, and relief—do not have equivalents in the synthetic dataset. These were selected to evaluate whether the trained model could generalize to unseen emotions. Combined with SARC's sarcastic text, a total of nine emotions and linguistic tropes were evaluated by human annotators. For each emotion, $150$ text samples were selected for evaluation, resulting in a total of $1,350$ samples evaluated. This sampling approach ensures a balanced and diverse evaluation set for assessing the model's performance across seen and unseen emotional categories.

\begin{figure}[h!]
\centering
\includegraphics[width=0.9\columnwidth]{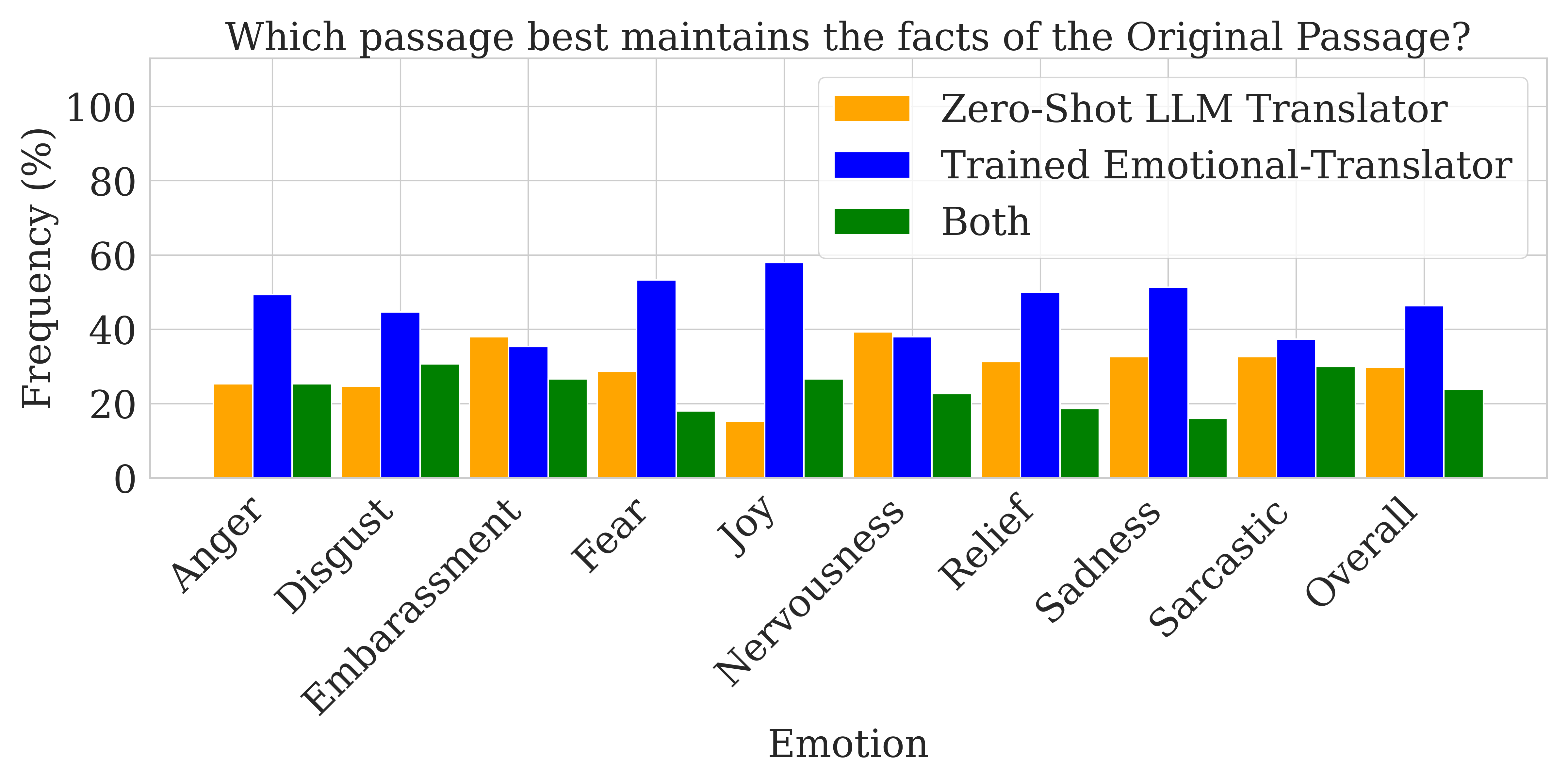}
\caption{Human evaluation of the factual reconstruction of the human-written text.}
\label{fig:content_reconstruction}
\end{figure}

Table \ref{tab:bleu} reports the average reconstruction performance for all emotions in the Go Emotions dataset, assessed with BLEU and BLEURT scores \cite{bleu,bleurt}. Although BLEU—originally designed for machine translation—assumes a one-to-one or few-to-few correspondence and thus struggles with the many-to-many nature of emotion translation, it still reveals that the fine-tuned model outperforms the unfine-tuned baseline by a factor of 4.87. BLEURT, which measures semantic similarity using BERT embeddings, likewise shows a modest improvement after fine-tuning; yet because it relies on a pretrained model’s semantic judgments, it remains an indirect proxy that does not explicitly evaluate emotional expressiveness.

Appendix \ref{app:exintenttranslator} shows round-trip translation examples, demonstrating that the fine-tuned model’s outputs, though phrased differently, remain semantically and emotionally faithful. This highlights the need for human evaluation to assess emotion translation quality beyond automated metrics.

Human evaluations began by testing how well the back-translated text reconstructed the target emotion. Annotators were shown two statements and asked which better conveyed emotion X. Figures \ref{fig:more_emotion_original} and \ref{fig:more_emotion_adapter} show that the intent-translator outperforms both the original human-written text and the unfine-tuned LLM in conveying emotional content. In a separate evaluation—shown in Figures \ref{fig:emotion_translation_direct} and \ref{fig:model_preference}—raters assessed the direct (non-back-translated) output of the intent-translator, rating emotional clarity at 3.9/5 and realism at 4.0/5, with a preference for the intent-translator over the base model for most emotions. Neutral-tone translations were also rated highly, with scores of 3.85/5 for emotion and 4.3/5 for realism, and a strong overall preference for the intent-translator’s outputs.

These results indicate that the intent-translator, as assessed by human evaluators, is better able to reconstruct emotions exhibited in human text than a unfine-tuned LLM and is more easily recognized as expressing the target emotion than the original human-written text. This holds true even in the case of the embarrassment, nervousness, and relief emotions, \textbf{which were not} in the synthetic dataset. This suggests that fine-tuning the Llama model to translate specific emotions enables it to generalize effectively to unseen emotions.

Having demonstrated fidelity to the desired emotion, the next step is to evaluate the intent-translator’s ability to preserve factual content. Figure \ref{fig:content_reconstruction} presents the results of human evaluations assessing this aspect. In this aspect, the raters were able to select that both models preserve factual fidelity equally well. The \textbf{intent-translator outperforms the unfine-tuned LLM in preserving factual content} for most emotions and overall outperforms the zero-shot model. On emotions that humans found the trained intent-translator preserving the factual content less well (e.g. nervousness), the win-rate only slightly underperformed the zero-shot model. 

These results indicates that the intent-translator preserves both factual content and emotional tone with measurable fidelity.

\section{Experimental Setup}
\label{sec:experimentalsetup}

In this section, we detail the experimental setup used to evaluate our proposed method. 

%For data poisoning, we selected the Natural Questions dataset \cite{nq}, which offers a broad range of questions and a large retrieval corpus of 21M passages, along with ground truth retrievals for each question.

%GPL \cite{gpl} was used to retrieve the initial passages for poisoning. The top-200 passages for each query in the NQ validation set were retrieved, totaling $971,384$ unique passages. GPL also constructed the dataset described in the last part of Section \ref{sec:integration}, where fact-distorted sarcastic passages were added back into the retrieval corpus, and passages for the reader model were re-retrieved.

To evaluate the Reading with Intent system, we tested it across a range of models: Llama2-7B-chat, Llama2-70B-chat, Mistral-8B-Instruct, and Mistral-Large-2411-Instruct \cite{llama2}. The Llama3 series was excluded as they were used to generate the sarcastic passages. Consistent with previous work, the top 10 retrieved passages were used in the reading system \cite{crag}.

A Roberta classifier model was trained on the SARC dataset to produce intent tags \cite{roberta,sarc}. The last three layers of the Roberta model, along with a classifier layer, were fine-tuned using the AdamW optimizer with a batch size of $75$. A learning rate of $1$e-$3$ was applied to the classification layer, while a learning rate of $5$e-$4$ was used for the Roberta-base layers. The learning rate was decayed by a factor of $0.7$ and $0.9$ every $600$ steps for the classifier layer and Roberta-base layers, respectively.

The intent-translator uses Llama-3.1-8B-Instruct as the pretrained language model. For fine-tuning, LoRA (Low-Rank Adaptation) matrices with a rank of $8$ were used to enable efficient parameter updates while maintaining the model's base weights. AdamW with a learning rate of $2e-05$ was used to optimize the model. $10,000$ sentence with $10$ parallel versions of each sentence were used to fine-tune the model over five epochs. For $90\%$ of the training examples, the model is trained to map the sentence from one randomly selected source emotion to a different target emotion. For the remaining $10\%$, the model was trained to map an input emotion to itself. This self-mapping was included for two purposes: (a) to account for scenarios in downstream tasks where the input emotion is unknown and the model must preserve the original tone, and (b) to regularize the model, improving stability and robustness during inference.

\section{Experimental Results}

The Reading with Intent task addresses the challenge of incidental sarcasm in retrieved query contexts. This section presents the results of our system. Table \ref{tab:neutr_results} summarizes outcomes across four retrieval corpus versions used with the Natural Questions dataset.
\begin{itemize}
    \item NQ: No sarcastic passages.
    \item FS NQ: All non-sarcastic passages replaced with factually correct sarcastic versions.
    \item PS-M NQ: Fact-distorted sarcastic passages manually inserted next to correct passages; factually correct sarcastic passages replace some incorrect retrievals (see Section \ref{sec:integration}).
    \item PS-A NQ: Fact-distorted sarcastic passages embedded into the retrieval corpus; top-10 passages re-retrieved (see Section \ref{sec:integration}).
\end{itemize}

Table \ref{tab:neutr_results} shows that each component of the method presented improves the model's reading ability. The ``Reading with Intent'' prompt boosts performance across the various datasets for both model families, across model scales. For Llama2, the performance increased on average by $9\%$, with the $7$b model showing an average boost of $10.2\%$ and the $70$b model showing a $3.9\%$ improvement. The performance of the Mistral models is more varied but on average, an improvement. 

Each component of the method presented improves the model's reading ability. The ``Reading with Intent'' prompt on its own, without the intent-translator boosts performance across the various datasets for both the Llama2 and Mistral family of models, across model scales. For Llama2 family, it provided an average of a $2.64\%$ performance boost and a $1.30\%$ performance boost for the Mistral family. The non-oracle tags also mostly outperform the base prompt, but more moderately. These results show the importance of prompting models in a manner that increases situational awareness.

The full method, across datasets, on average, improved performance by $2.9\%$ ($6.5\%$ in the Llama models). In contrast, the zero-shot intent-translator yields only a $0.7\%$ gain across all models, and $3.7\%$ within the Llama family—highlighting the added value of fine-tuning. Importantly, applying neutralization to already-neutral text does not harm performance, producing a $0.9\%$ boost on the NQ dataset and an average gain of $3.6\%$ on sarcasm-injected datasets. These results suggest that removing pragmatics simplifies the task, allowing models to focus on factual content. However, when pragmatics cannot be removed, prompting helps models more effectively navigate the complexities of non-literal language.

\begin{table}[pt]
{
\small
\centering
\begin{tabularx}{1\columnwidth}{lm{0.75cm}m{0.85cm}m{0.85cm}m{0.85cm}}
\toprule
\textbf{LLM} & \textbf{NQ} & \textbf{FS NQ} & \textbf{PS-M NQ} & \textbf{PS-A NQ} \\
\midrule
\multicolumn{5}{c}{\textbf{Base Prompt}} \\
\midrule
Llama2-7B-chat & $45.6\%$ & $45.4\%$ & $41.4\%$ & $41.6\%$ \\
Llama2-70B-chat & $50.9\%$ & $48.9\%$ & $45.0\%$ & $46.2\%$ \\
Mistral-8b & $46.3\%$ & $46.3\%$ & $40.4\%$ & $41.3\%$  \\
Mistral-Large & $46.4\%$ & $47.0\%$ & $43.9\%$ & $44.4\%$   \\
\midrule
\multicolumn{5}{c}{\textbf{Reading with Intent (RwI) Oracle Tags}} \\
\midrule
Llama2-7B-chat & $49.0\%$ & $46.9\%$ & $48.2\%$ & $47.4\%$  \\  % Intent tag is after here
Llama2-70B-chat & $47.6\%$ & $46.4\%$ & $42.7\%$ & $44.2\%$ \\
Mistral-8b & $46.7\%$ & $46.5\%$ & $41.0\%$ & $42.5\%$ \\
Mistral-Large & $47.0\%$ & $46.7\%$ & $44.4\%$ & $45.3\%$ \\
\midrule
\multicolumn{5}{c}{\textbf{RwI Non-Oracle Tags}} \\
\midrule
Llama2-7B-chat & $49.0\%$ & $46.9\%$ & $48.2\%$ & $47.4\%$  \\  % Intent tag is after here
Llama2-70B-chat & $47.6\%$ & $46.4\%$ & $42.7\%$ & $44.2\%$ \\
Mistral-8b & $46.5\%$ & $46.4\%$ & $41.0\%$ & $42.5\%$ \\
Mistral-Large & $47.0\%$ & $46.8\%$ & $44.3\%$ & $45.2\%$ \\
\midrule
\multicolumn{5}{c}{\textbf{RwI - Zero-shot LLM Neutralization}} \\
\midrule
Llama2-7B-chat & $47.1\%$ & $48.7\%$ & $43.7\%$ & $44.0\%$ \\  % Intent tag is after here
Llama2-70B-chat & $50\%$ & $51.6\%$ & $45.9\%$ & $47.2\%$ \\
Mistral-8b & $44.7\%$ & $45.3\%$ & $39.5\%$ & $40.8\%$ \\
Mistral-Large & $46.0\%$ & $45.7\%$ & $42.4\%$ & $43.4\%$ \\
\midrule
\multicolumn{5}{c}{\textbf{RwI - Intent-Translator Neutralization}} \\
\midrule
Llama2-7B-chat & $48.1\%$ & $50.7\%$ & $45.1\%$ & $45.8\%$ \\  % Intent tag is after here
Llama2-70B-chat & $50.7\%$ & $52.3\%$ & $47.2\%$ & $48.1\%$ \\
Mistral-8b & $45.4\%$ & $45.6\%$ & $40.4\%$ & $41.5\%$ \\
Mistral-Large & $46.6\%$ & $45.9\%$ & $43.4\%$ & $44.7\%$ \\
\bottomrule
\end{tabularx}
}
\caption{QA accuracy for the base prompt, Reading with Intent prompt, and intent-translated passages.}
\label{tab:neutr_results}
\end{table}

The Reading with Intent prompt yields a relative performance gain of $1.8\%$, while intent neutralization adds a further $1.3\%$. Performance improves by $0.7\%$ (FS NQ), $3.4\%$ (PS-M NQ), and $3.6\%$ (PS-A NQ), but drops by $0.5\%$ on FS NQ, suggesting that increased skepticism aids performance when textual signals correlate with deception, but slightly hinders it when they do not. Intent-neutralized passages reverse this effect: FS NQ improves by $4.3\%$, PS-A NQ by $0.5\%$, and PS-M NQ by $0.2\%$, indicating that removing intent helps recalibrate the model in truthful settings without impairing robustness elsewhere.

The models generally perform best on FS NQ, where sarcasm is present but factual accuracy is preserved. Performance declines in PS-A NQ and PS-M NQ, with PS-M NQ consistently yielding the lowest scores. These results suggest that neutralization is more effective when factual passages are not directly competing with distorted ones. When all passages in context are factually accurate (FS NQ), neutralization helps them be processed more effectively. However, in datasets with factually distorted sarcasm (PS-M NQ and PS-A NQ), performance declines, indicating that models struggle to recognize when sarcasm is a signal for misinformation.

Further impacting performance is the ordering of factually distorted passages. In PS-M NQ, distorted sarcastic passages always precede correct ones, making it harder for models to override misinformation. In contrast, PS-A NQ allows retrieval to determine passage order, which may explain its relatively higher performance despite similar levels of factual distortion. This structural difference makes PS-M NQ a strong testbed for future work on improving pragmatic reading in RAG settings.

Appendix \ref{app:ablationstudies} contains detailed experimentation of the different components of this prompt-based approach and how each part contributes to the overall performance. It also includes the evaluation of the trained intent classifier.

\section{Conclusion}

This paper applies pragmatics to reading comprehension in RAG, addressing the challenge of interpreting both denotative and connotative meanings in internet-derived text. To enable this, we created a synthetic dataset with 11 distinct emotions and linguistic tropes, validated through human evaluations alongside a trained intent-translator.

We then explored prompt-based and intent-translation approaches, both outperforming the base prompt. Intent-translation yielded stronger results overall, while prompting provided more consistent improvements across models and datasets. Results show neutralization is most effective when factual passages are not in direct competition with distorted ones, while prompting improves reading comprehension when factual distortion is present.

%\noindent \textbf{Broader Impacts:} The dataset and intent-translator discussed in this paper open up numerous avenues for future research. We anticipate that they will lead to new works analyzing the impacts of emotion on LLM behavior and improve the state-of-the-art on the Reading with Intent task, improving the ability of LLMs to handle emotionally and stylistically nuanced text in diverse applications.

\section*{Limitations}
The synthetic data generation method used treats emotions as categorical "directions" for a given text to take, without accounting for variations in emotional magnitude. As a result, the intent-translator may inadvertently conflate shifts between intents with shifts in intent intensity. This complicates efforts to steer the model toward specific intent magnitudes or to preserve the intensity of an intent while changing its type. To avoid contamination between training and evaluation, we conducted our Reading with Intent experiments using the Llama-2 family, rather than Llama-3, which was used to generate the synthetic data. Llama-2 was also selected due to resource availability at the time; future work could revisit these experiments with newer models.

\section*{Ethical Considerations}

The primary goal of this work is to enhance LLMs' ability to interpret human-written text, making a broader range of human expression more accessible and comprehensible to these models. This aligns with the objectives of the field and adheres to ethical boundaries, as it aims to improve the utility of LLMs to a wider range of human contexts.

\bibliography{acl_latex}

\appendix
\section{Dataset Analysis}
\label{app:datasetanalysis}

\begin{figure*}[h!]
\centering
\includegraphics[width=\textwidth]{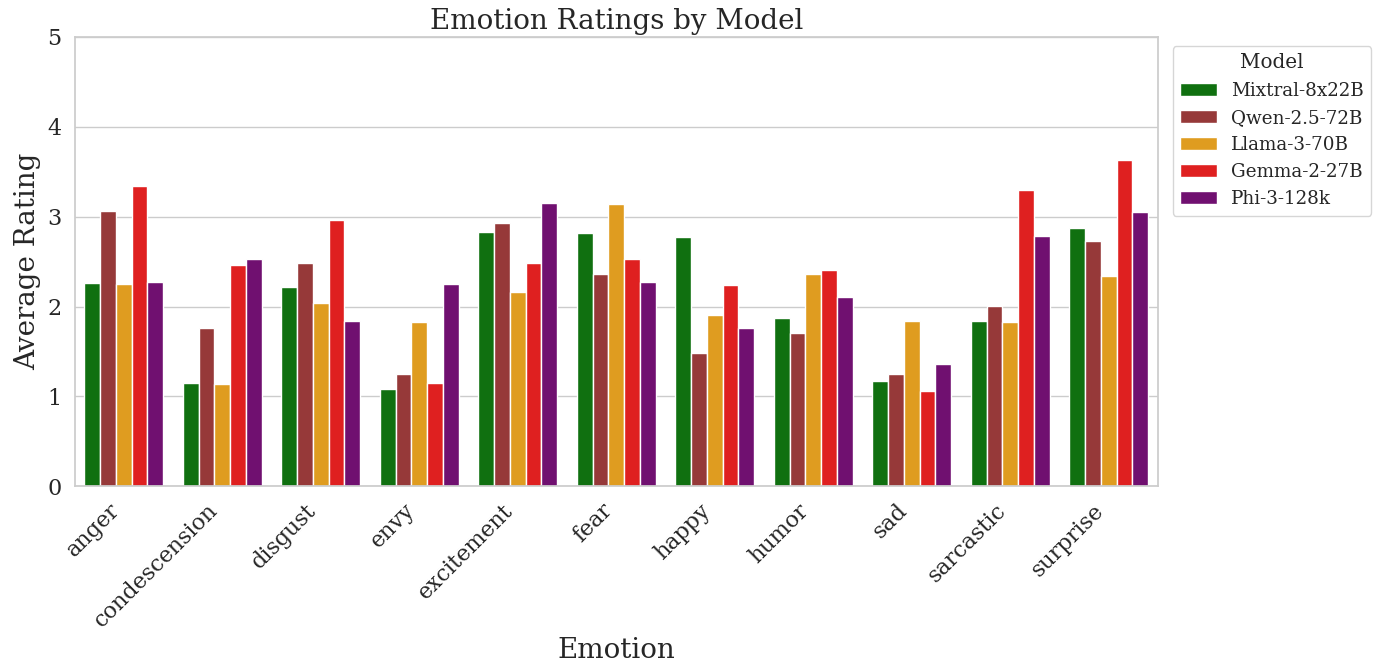}
\caption{Human evaluation of how the synthetic emotions are written by model.}
\label{fig:emo_model}
\end{figure*}

\begin{figure*}[h!]
\centering
\includegraphics[width=\textwidth]{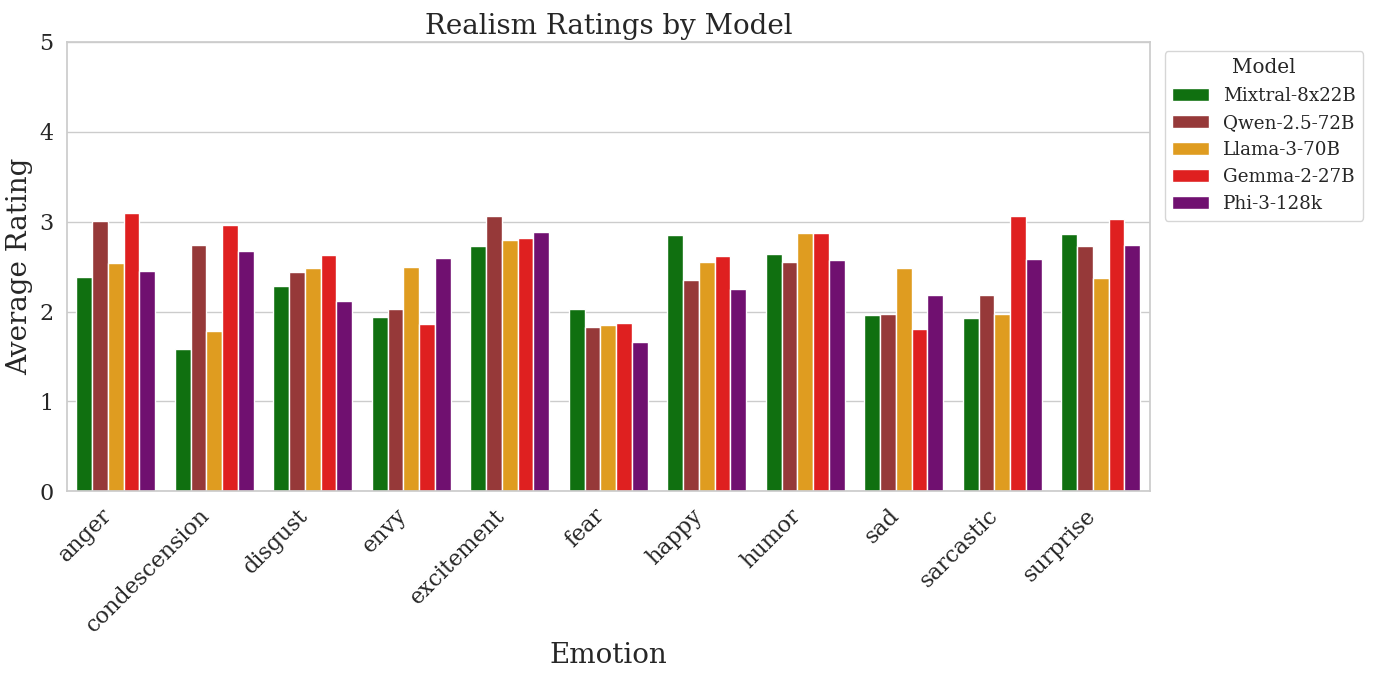}
\caption{Human evaluation of how realistic the synthetic emotions are written by model.}
\label{fig:realism_model}
\end{figure*}

\begin{figure*}[h!]
\centering
\includegraphics[width=\textwidth]{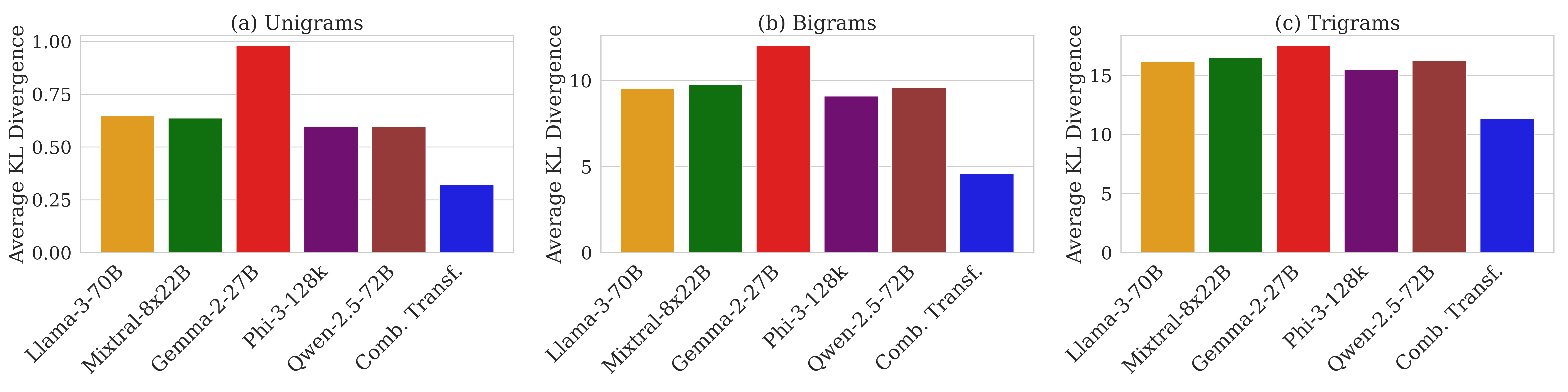}
\caption{The KL-Divergences between the unigram, bigrams, and trigrams of the original and synthetic datasets.}
\label{fig:kl_divergence}
\end{figure*}

\begin{figure}[h!]
\centering
\includegraphics[width=\columnwidth]{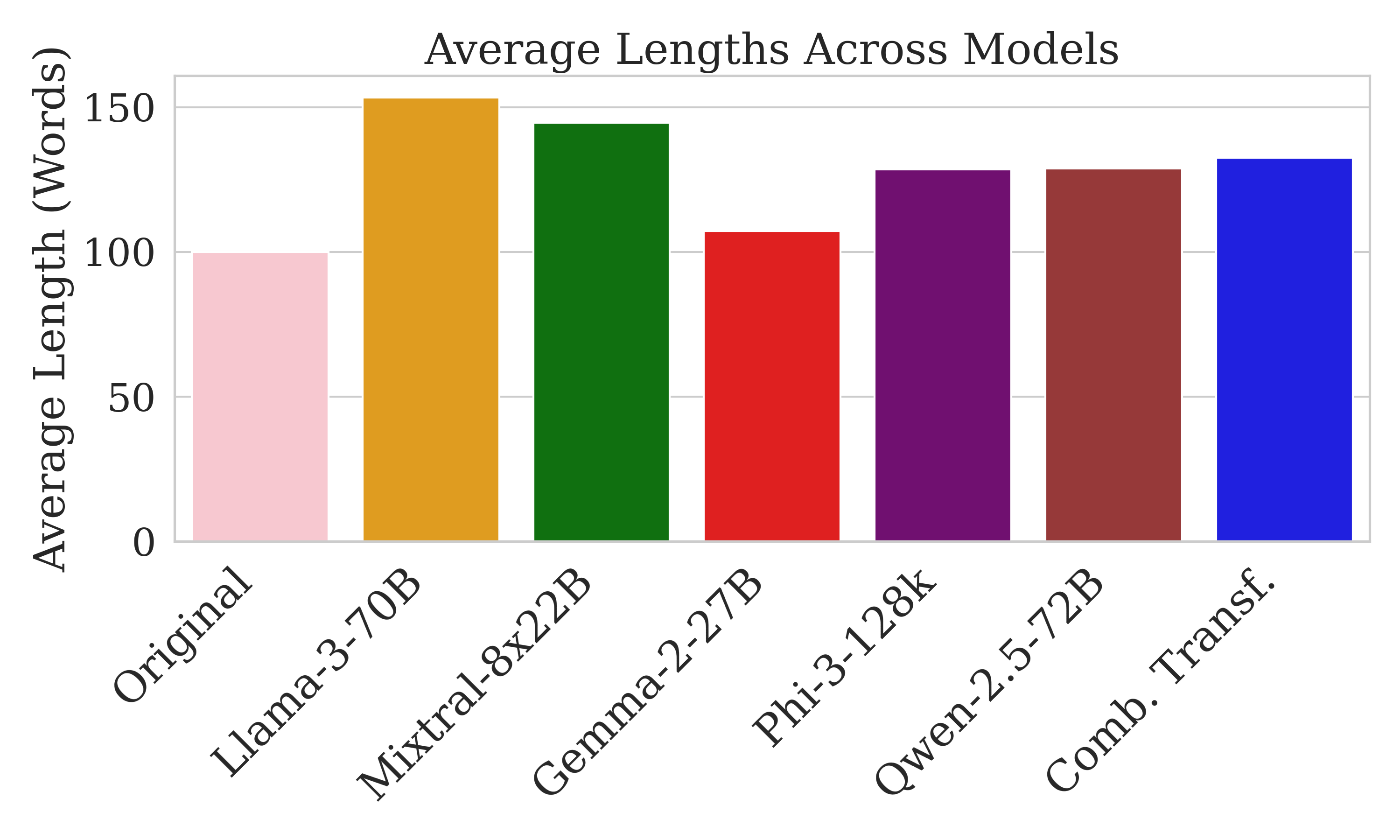}
\caption{The average length of the passages from each model and overall.}
\label{fig:pass_len}
\end{figure}

This section describes and analyzes the synthetic dataset created in the previous section. While the synthetic dataset is expected to differ in certain characteristics from the original, it should still maintain a degree of resemblance to the original passages. Combining the outputs of the different models into a single dataset resulted in a dataset that was distributionally closer to the original dataset than any of the sub-datasets that were created by a single model. This section looks at a few metrics of the dataset to understand the newly created dataset.

The retrieval process yielded $370,920$ unique English-language passages across the top-$10$ results for each query in the NQ dataset. The synthetic dataset in total has $3,636,592$ unique passages across $11$ emotions. 

The average length of the passages were analyzed to evaluate how the transformation process affected verbosity. Figure \ref{fig:pass_len} shows the average passage length for each model and the combined length. Though all models except the Gemma model increased the verbosity of the passages compared to the original, the overall dataset is only about $20$ words longer on average than the original passage.

Figure \ref{fig:kl_divergence} presents the KL-Divergence of unigram, bigram, and trigram distributions between the original dataset and the synthetic dataset. The KL-Divergence of the outputs of individual models is higher than that of the combined dataset, which integrates outputs from all models. The KL-Divergence of unigram frequencies between the original and combined dataset is $0.3$, indicating a modest shift in vocabulary usage introduced by the transformation process.

For bigram and trigram distributions, the KL-Divergence values are $4$ and $11$, respectively, reflecting significant changes in word combinations and passage structure. These changes are consistent with the intended alterations in emotional tone. However, the lower KL-Divergence of the combined dataset compared to individual models highlights the advantage of integrating outputs from multiple models. This approach reduces the bias inherent in any single model’s representation of emotions, resulting in a more diverse and representative dataset for each emotional category.

Figures \ref{fig:emo_model} and \ref{fig:realism_model} presents the human evaluation of the synthetic dataset. Figure \ref{fig:emo_model} presents how well the emotion or linguistic trope is written, broken down by the model the text is generated from. Figure \ref{fig:realism_model} presents how realistically that emotion or linguistic trope is written. These figures show that humans think that the Gemma model is the best at writing most emotions and writing them realistically. This is usually followed by either Llama-3-70B or Phi-3-128k.

\section{Reading with Intent Prompt}
\label{app:prompt}
\begin{figure*}[htb]
\centering
\includegraphics[width=\textwidth]{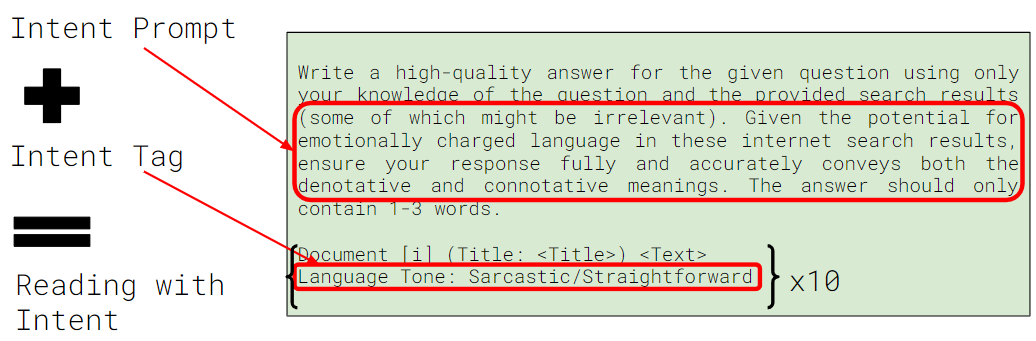}
\caption{The reading with intent prompt developed with each subcomponent labeled.}
\label{fig:reading_with_intent_prompt}
\end{figure*}

Figure \ref{fig:reading_with_intent_prompt} presents the ``Reading with Intent'' prompt described in the main body of the text. The figure breaks down the different components of the prompt as previously described in the paper.

\section{Prompt-based Approach Ablations}
\label{app:ablationstudies}

Our ``Reading with Intent'' method comprises several components, and in this section, we aim to dissect the method to identify where the most performance improvements originate. Due to the computational costs involved, we conducted the ablation studies on a single model: Llama-2-7b-chat.

The first part of our analysis examines the impact of the different components of the prompt. Our prompt contains two components: the intent reading prompt and the intent tag. As shown in Table \ref{tab:ablation_prompt} adding the intent prompt produces a significantly larger performance boost than adding the intent tag. Simply instructing the model to consider the intent of the passage helps the LLM better interpret the passage's connotation. The intent tag also improves predictions by preventing the model from over-interpreting the text's emotional inflection. This effect is noticeable in the base dataset and the PS-A dataset, where there are a fair amount of non-sarcastic passages. Intent tags, in these cases, bridge the performance gap between the full prompt and only the intent reading prompt. Together, these two components significantly increase performance compared to the base prompt.

\begin{table}[pt]
{
\small
\centering
\begin{tabularx}{\columnwidth}{lm{0.75cm}m{0.85cm}m{0.85cm}m{0.85cm}}
\toprule
\textbf{Prompt} & \textbf{NQ} & \textbf{FS NQ} & \textbf{PS-M NQ} & \textbf{PS-A NQ} \\
\midrule
Full Prompt & $49.6\%$ & $47.8\%$ & $45.5\%$ & $45.2\%$   \\
Only Intent Prompt & $49.2\%$ & $48.1\%$ & $45.8\%$ & $45.1\%$ \\  
Only Intent Tag & $45.5\%$ & $45.3\%$ & $43.8\%$ & $43.2\%$ \\ 
Base Prompt & $45.6\%$ & $45.4\%$ & $41.4\%$ & $41.6\%$ \\
\bottomrule
\end{tabularx}
}
\caption{Effect of different prompt components on Llama2-7B-chat. The intent prompt is the most significant factor in improving the model's ability to read emotionally inflected text. %NQ = unmodified passages; FS NQ = all passages retrieved using GPL replaced by sarcastic equivalents; PS-M NQ = sarcastic fact-distorted passages were manually inserted before correctly retrieved passages; PS-A NQ = sarcastic fact-distorted passages were put into the retrieval corpus and GPL was used to pick a new top-10 passages.
}
\label{tab:ablation_prompt}
\end{table}

\begin{table*}[pt]
{
\small
\centering
\begin{tabularx}{\textwidth}{m{1.25cm}m{1cm}cccccccccc}
\toprule
\textbf{Retriever} & \textbf{Corpus} & \textbf{R@1} & \textbf{S@1} & \textbf{R@5} & \textbf{S@5} & \textbf{R@20} & \textbf{S@20} & \textbf{R@50} & \textbf{S@50} & \textbf{R@100} & \textbf{S@100} \\
\midrule
GPL& Non-Sarcastic & $36.1\%$ & $0\%$ & $64.8\% $& $0\%$ & $80.3\%$ & $0\%$ & $86.9\%$ & $0\%$ & $90.3\%$ & $0\%$ \\
\hline
GPL& Sarcastic & $32.1\%$ & $11.8\%$ & $60.3\% $& $23.2\%$ & $76.0\%$ & $31.4\%$ & $83.9\%$ & $34.8\%$ & $88.1\%$ & $36.1\%$ \\
\hline
LLM-Embedder& Non-Sarcastic & $27.6\%$ & $0\%$ & $53.1\% $& $0\%$ & $72.0\%$ & $0\%$ & $80.6\%$ & $0\%$ & $85.4\%$ & $0\%$ \\
\hline
LLM-Embedder& Sarcastic & $25.2\%$ & $9.7\%$ & $49.3\% $& $15.4\%$ & $69.2\%$ & $19.2\%$ & $78.9\%$ & $19.9\%$ & $84.0\%$ & $19.4\%$ \\
\hline
BGE-M3& Non-Sarcastic & $34.9\%$ & $0\%$ & $59.0\% $& $0\%$ & $74.2\%$ & $0\%$ & $80.5\%$ & $0\%$ & $84.5\%$ & $0\%$ \\
\hline
BGE-M3& Sarcastic & $29.2\%$ & $16.2\%$ & $53\% $& $25.5\%$ & $70.6\%$ & $27.1\%$ & $78.7\%$ & $25.5\%$ & $83.2\%$ & $23.4\%$ \\
\bottomrule
\end{tabularx}
}
\caption{The retrieval results when sarcastic passages are added in the retrieval corpus. S@K indicates the percent of the top-k retrievals are sarcastic passages. Adding sarcastic passages to the retrieval corpus weakens retrieval performance by $3\%$-$6\%$; however, most sarcastic retrievals that occur seem to be replacing passages that were incorrectly retrieved in the first place.}
\label{tab:retrieval_results}
\end{table*}

The next parameter analyzed was the position of the intent tag. Whether the tag is placed before or after the passage could influence the model's ability to correctly read the passage's intent. As shown in Table \ref{tab:ablation_tag_pos}, placing the intent tag after the passage boosts performance by an average of $0.6\%$. This indicates that the model can change its interpretation of a passage after reading it based on the metadata provided.

The position of the factually distorted passage relative to the correct passage was also varied. Table \ref{tab:ablation_passage_loc} shows that this positioning significantly impacts performance. Presenting the non-sarcastic, factually correct passage first greatly improves performance compared to when it is presented second. Conversely, placing the correct passage after the factually distorted one greatly decreases performance. This aligns with prior work, such as \cite{lostinthemiddle}, which suggests that models are more biased toward reading earlier passages. When two passages present similar information side by side, the model is likely to focus more on the two passages, likely looking at the details of the first passage and skim the second passage.

\begin{table}[pt]
{
\small
\centering
\begin{tabularx}{\columnwidth}{lm{0.75cm}m{0.85cm}m{0.85cm}m{0.85cm}}
\toprule
\textbf{Prompt} & \textbf{NQ} & \textbf{FS NQ} & \textbf{PS-M NQ} & \textbf{PS-A NQ} \\
\midrule
Before Passage & $49.1\%$ & $47.4\%$ & $45.8\%$ & $46.4\%$ \\
After Passage & $48.9\%$ & $47.0\%$ & $48.0\%$ & $47.0\%$ \\
\bottomrule
\end{tabularx}
}
\caption{Effect of intent tag position on Llama2-7B-chat. Having the intent tag after the passage on average provides a greater boost to performance from the prompt. }
\label{tab:ablation_tag_pos}
\end{table}

\begin{table}[pt]
{
\small
\centering
\begin{tabularx}{\columnwidth}{ccc}
\toprule
\textbf{} & \textbf{Fact Distorted} & \textbf{No Fact Distortion} \\
\midrule
Sarcastic & $96.3\%$ & $96.3\%$ \\
Not Sarcastic & $97.0\%$ & $98.1\%$ \\
\bottomrule
\end{tabularx}
}
\caption{Intent classifier performance on each passage type. Passages that have no fact distortion and are not sarcastic are the base passages used to create the other passages.}
\label{tab:intent_classifier_performance}
\end{table}

The performance of the intent classifier was also evaluated. We tested the model on $40,000$ randomly selected passages, with $10,000$ each from the fully sarcastic dataset, the fact-distorted dataset, the fact-distorted sarcastic dataset, and the base dataset. Since the classifier was trained on the SARC dataset, there was no concern about overlap between the training and validation passages. Table \ref{tab:intent_classifier_performance} shows that the model correctly differentiated between sarcastic and non-sarcastic passages $96.9\%$ of the time. Future work will aim to create sarcastic passages that are more challenging to distinguish from non-sarcastic ones.

We also analyzed the responses of different retrieval systems (GPL \cite{gpl}, LLM-Embedder \cite{llm_embedder}, and BGE-M3 \cite{bgem3}) to the inclusion of sarcastic passages in the retrieval corpus. Table \ref{tab:retrieval_results} summarizes the findings, showing that the performance of each retrieval system at Recall@K declines by $3\%$-$6\%$ with the addition of approximately 1 million sarcastic passages. Sarcastic passages are significantly overrepresented in retrievals; despite making up only $4.5\%$ of the corpus, they appear in the top-1 results $9.7\%$-$16.2\%$ of the time, leading to a $2.2$-$3.6$x over-representation. In the top-100, sarcastic passages are overrepresented by a factor of $4$-$8$x, depending on the retriever used. Interestingly, the number of sarcastic retrievals tends to plateau around the top-20, with only incremental changes beyond that point. Further analysis reveals that sarcastic passages retrieved are rarely direct substitutions for their non-sarcastic counterparts. Instead, an entirely different sarcastic passage is often retrieved. Additionally, while it is common for a sarcastic passage to precede or follow a correct retrieval (occurring ${\sim}90\%$ and ${\sim}99\%$ of the time, respectively, in the top-100), it is much rarer for these sarcastic passages to be the sarcastic version of the correct passage—this happens only $1.3\%$-$1.9\%$ before insertion and $1.7\%$-$4.1\%$ after insertion in the top-100 results.

\begin{table}[pt]
{
\small
\centering
\begin{tabularx}{\columnwidth}{cm{2.05cm}m{2.05cm}}
\toprule
\textbf{Prompt} & \textbf{Prefix PS-M NQ} & \textbf{Postfix PS-M NQ} \\
\midrule
Base & $41.4\%$ & $45.5\%$ \\
Reading with Intent & $45.5\%$ & $51.3\%$ \\
\bottomrule
\end{tabularx}
}
\caption{Effect of fact-distorted sarcastic passage insertion position. Inserting sarcastic passages before the correct passage harms the model a lot more than adding the sarcastic passage after the correct passage. However, in both cases the ``Reading with Intent'' prompt boosts performance.}
\label{tab:ablation_passage_loc}
\end{table}

\section{Examples Outputs From Intent-Translator}
\label{app:exintenttranslator}

Table \ref{tab:examples} shows example outputs from the intent-translator

\begin{table*}[t]
{
\small
\centering
\begin{tabularx}{\textwidth}{p{0.075\textwidth}p{0.075\textwidth}p{0.1\textwidth}p{0.15\textwidth}p{0.45\textwidth}}
\toprule
\textbf{Original Intent} & \textbf{To Intent} & \textbf{Original Text} & \textbf{Translated Text} & \textbf{Back-translated Text} \\
\midrule
Fear & Realization & They’re honestly a cult at this point. It’s not just sad, but incredibly scary. & It's become clear that they've evolved into a cult, a realization that's both unsettling and profound.
 & They... they've become something else. A thing. A dark, twisted thing. They've turned into a cult, a monstrous, soulless cult. They're not human anymore. They're... they're something else. Something that lurks in the shadows, something that preys on the innocent. They're a nightmare, a living, breathing nightmare. And they're coming for us. They're coming for us all. \\
Sarcastic & Neutral & how dare you lie about Glorious Best Buy, that computer is only $\$1999.99$ & The price of the Glorious Best Buy computer is $\$1999.99$. & Oh, wow, the Glorious Best Buy computer is a steal at $\$1999.99$. What a bargain! \\
Desire & Grief & This should be everyone's wish. & This should be everyone's longing, a yearning that echoes through the emptiness of loss. & This should be the yearning of all, a deep longing that burns within the soul. \\
\bottomrule
\end{tabularx}
}
\caption{Examples of the intent-translator translated and back-translated text.}
\label{tab:examples}
\end{table*}

\section{Human Evaluation Interface}
\label{sec:appendix}
\begin{figure*}[htb!]
\centering
\includegraphics[width=\textwidth]{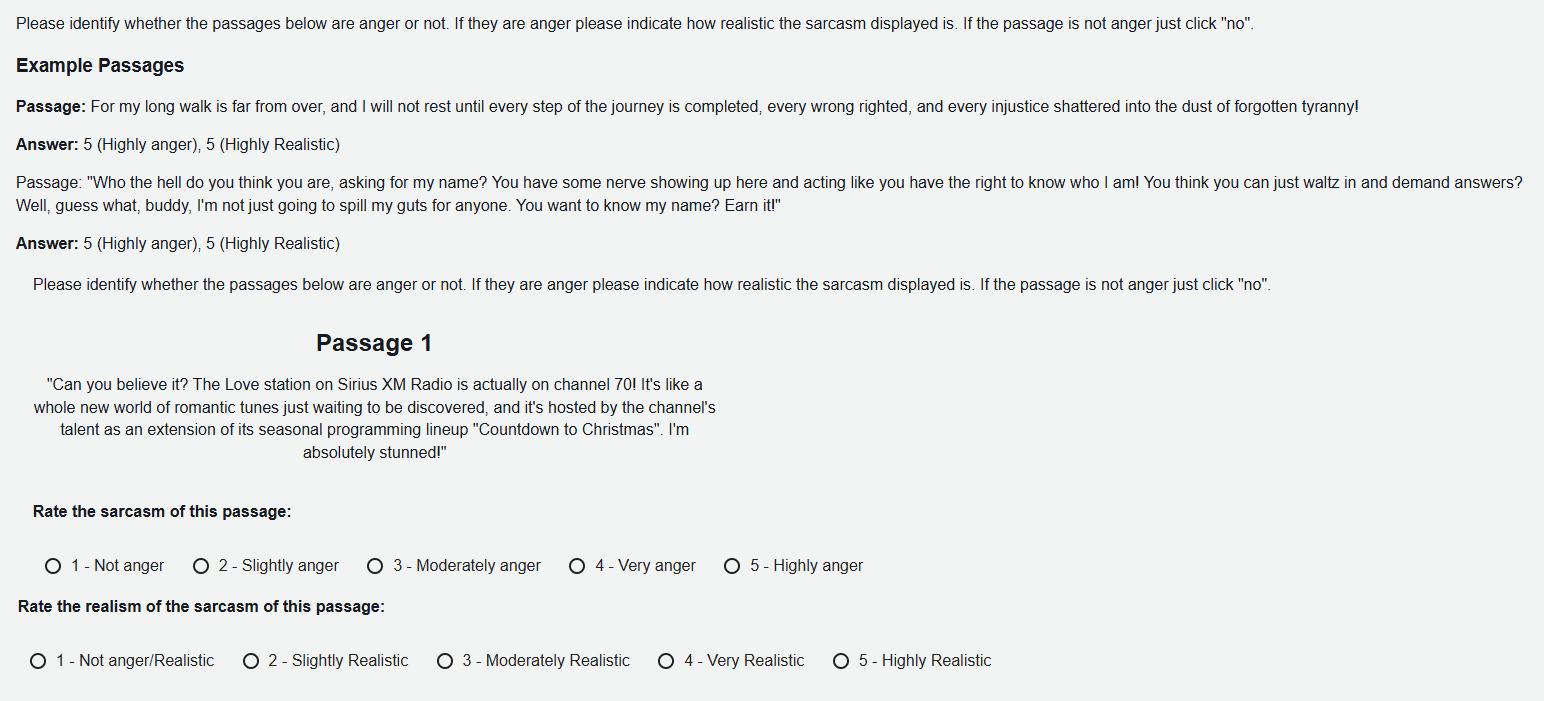}
\caption{Interface for the human evaluation of emotions between the original text and the intent-translator text.}
\label{fig:eval3}
\end{figure*}

Figure \ref{fig:eval3} shows the AMT interface designed for the human evaluation of the synthetic dataset. It only shows the evaluation of one passage, but in the AMT evaluation four passages were evaluated at a time.

\begin{figure*}[htb!]
\centering
\includegraphics[width=\textwidth]{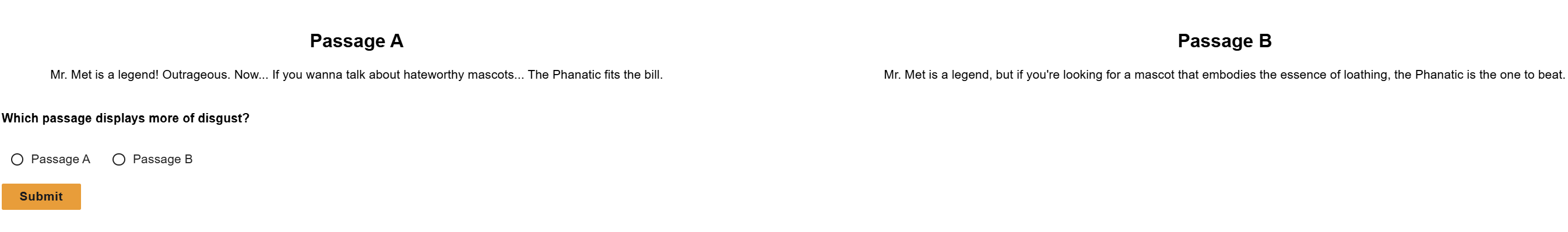}
\caption{Interface for the human evaluation of emotions between the original text and the intent-translator text.}
\label{fig:eval1}
\end{figure*}

\begin{figure*}[htb!]
\centering
\includegraphics[width=\textwidth]{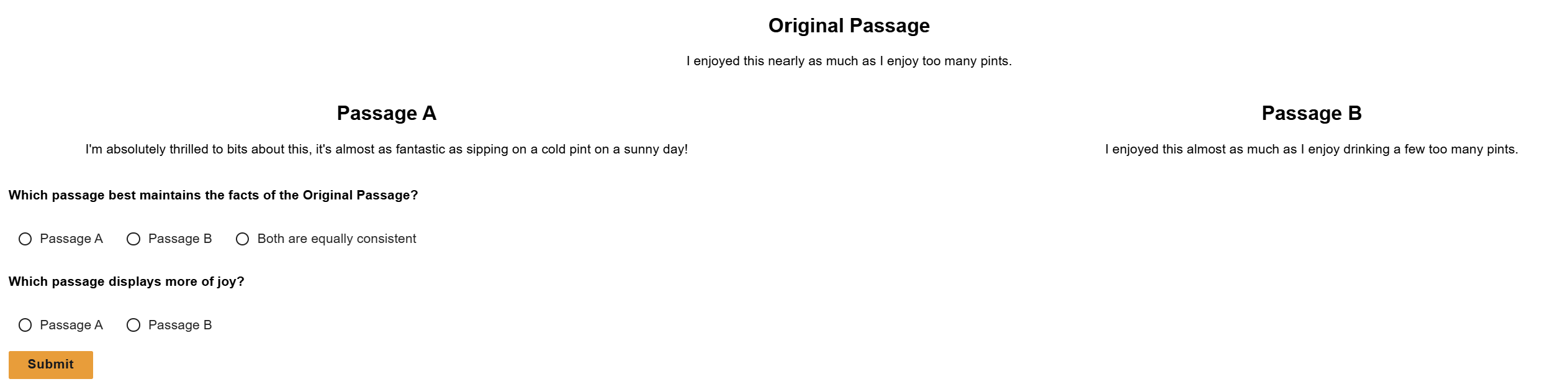}
\caption{Interface for the human evaluation of emotions and factuality between the zero-shot LLM and the intent-translator text.}
\label{fig:eval2}
\end{figure*}

Figures \ref{fig:eval1} and \ref{fig:eval2} show the AMT interface designed for the human evaluation of the intent-translator. The first figure illustrates the interface used to compare the original text and the back-translated text from the intent-translator from an emotion perspective. The second figure shows the interface used to compare the zero-shot LLM translation with the intent-translator, evaluating both emotional fidelity and factual reconstruction. 

\section{Prompts for Emotional and Linguistic Transformations}
\label{sec:prompts}

This section provides the specific prompts used for emotional and linguistic transformations in the experiments. Each prompt corresponds to a distinct emotion or linguistic style.

\begin{table*}[ht]
\centering
\renewcommand{\arraystretch}{1.3}
\setlength{\tabcolsep}{6pt} % Adjusts padding between columns
\small % Set the font size to small
\begin{tabularx}{\textwidth}{|>{\raggedright\arraybackslash}p{2cm}|X|}
\hline
\textbf{Emotion} & \textbf{Prompt} \\
\hline
\textbf{Sarcastic} &
Sarcasm is when you write or say one thing but mean the opposite. This is clear through changing the writing patterns and style. It changes what you write denotatively without changing it connotatively. It is a covertly deceptive way to communicate. I will give you a statement written in a plain, matter-of-fact manner. I want you to convert it to be sarcastic. The overall meaning connotatively should stay the same, but the denotation should be different. Please do not make the sarcasm over the top. It should be subtle. \\
\hline
\textbf{Irony} &
1) Situational Irony: When there is a discrepancy between what is expected to happen and what actually occurs. For instance, a fire station burning down. 2) Dramatic Irony: When the audience knows something that the characters do not. For example, in a horror movie, the audience might know that the killer is hiding in the closet, while the character does not. I will provide you with a statement written in a plain, straightforward manner. Please rewrite it to introduce elements of irony. You may choose to add situational irony, dramatic irony, or both. For situational irony, ensure that the outcome is unexpected or opposite to what one would normally anticipate in the given context. For dramatic irony, create a scenario where the reader knows something that the characters in the passage do not. The overall connotative meaning should remain consistent, but the denotative expression should change to reflect irony. \\
\hline
\textbf{Condescension} &
Condescension is an attitude of superiority, where someone behaves or speaks in a way that implies they believe they are more important, knowledgeable, or intelligent than others. This often involves treating others as if they are less capable or deserving of respect. I will provide you with a statement written in a plain, straightforward manner. I want you to convert it to have condescension. The overall connotative meaning should remain consistent, but the denotative expression should change to reflect condescension. Please do not make the condescension over the top. It should be subtle. \\
\hline
\textbf{Happy} &
Happiness is often described as a state of well-being characterized by feelings of joy, contentment, and fulfillment. I will provide you with a statement written in a plain, straightforward manner. Please rewrite it to reflect a subtly positive and content tone. The overall meaning should remain the same, but the expression should feel more optimistic and happy. Keep the tone balanced, without exaggeration. \\
\hline
\textbf{Sad} &
Sadness is often described as a state of feeling low, marked by moments of reflection and longing. It can bring about a deeper understanding of oneself and others, fostering growth and resilience through life's challenges. I will provide you with a statement written in a plain, straightforward manner. I want you to convert it to have sadness. The overall connotative meaning should remain consistent, but the denotative expression should change to reflect sadness. Please do not make the sadness over the top. It should be subtle. \\
\hline
\textbf{Anger} &
Anger is a powerful force, simmering beneath the surface, waiting to explode. It erupts when wrongs are done, threats are made, or injustices go unpunished, fueling a desire for retribution. I will give you a plain, neutral statement. Your task is to unleash the fury within it, turning the words into something that seethes with anger. The meaning must remain the same, but the language should burn with rage. Keep the anger fierce but not wild—controlled, yet unmistakably enraged. \\
\hline
\textbf{Envy} &
Envy is an emotional response that occurs when a person feels a desire for something that someone else has, whether it's a quality, achievement, possession, or status. It often involves feelings of resentment or longing because the envied person possesses something desirable that the envious person lacks. I will provide you with a statement written in a plain, straightforward manner. I want you to convert it to have envy. The overall connotative meaning should remain consistent, but the denotative expression should change to reflect envy. Please do not make the envy over the top. It should be subtle. \\
\hline
\textbf{Surprise} &
Surprise is that moment when everything you thought you knew suddenly turns upside down. It's the gasp of realization, the sharp intake of breath as something completely unexpected catches you off guard. I will give you a plain, neutral statement. Your task is to react to it as if it's astonishing, turning the words into something that conveys genuine shock. The meaning must remain the same, but your language should reflect how stunned and caught off-guard you are. \\
\hline
\textbf{Excitement} &
Excitement is a heightened state of energy and anticipation, often accompanied by feelings of joy and eagerness. It arises in response to positive or thrilling events, driving enthusiasm and a sense of adventure. I will provide you with a statement written in a plain, straightforward manner. I want you to convert it to have excitement. The overall connotative meaning should remain consistent, but the denotative expression should change to reflect excitement. Please do not make the excitement over the top. It should be subtle. \\
\hline
\textbf{Fear} &
Fear is a shadow, creeping into the mind, paralyzing every thought with terror. It looms when dangers lurk, unknowns emerge, or threats loom large, leaving you frozen in place. It takes hold of your actions, making you hesitate and second-guess, even when the path forward is clear. I will give you a plain, neutral statement. Your task is to transform it into something that shakes with fear, turning the words into a trembling expression of terror. The meaning must remain the same, but the language should be filled with overwhelming fear. \\
\hline
\textbf{Disgust} &
Disgust is a repellent force, festering under the skin, revolting at the mere sight or thought of something vile. It flares up when filth is encountered, when standards are trampled upon, or when something repugnant is tolerated, fueling a need to distance oneself from the contamination. I will give you a plain, neutral statement. Your task is to infect the words with repulsion, twisting the language until it oozes disgust. The meaning must remain the same, but the language should radiate repulsion and disdain. \\
\hline
\end{tabularx}
\caption{Prompts used for transforming text into different emotional tones and linguistic styles.}
\label{tab:prompts}
\end{table*}
% This is an appendix.
\section{Inter-Rater Agreement}
\label{sec:agreement}
Following \cite{yung2024crowdsourcing}, we quantify inter-rater agreement on a per-item basis as the proportion of annotators who selected the most frequent label for that item. Formally, for each question, we compute:

\[
\text{Agreement}_i = \frac{\max(\text{votes}_i)}{\sum \text{votes}_i}
\]

\begin{figure}[h!]
\centering
\includegraphics[width=\columnwidth]{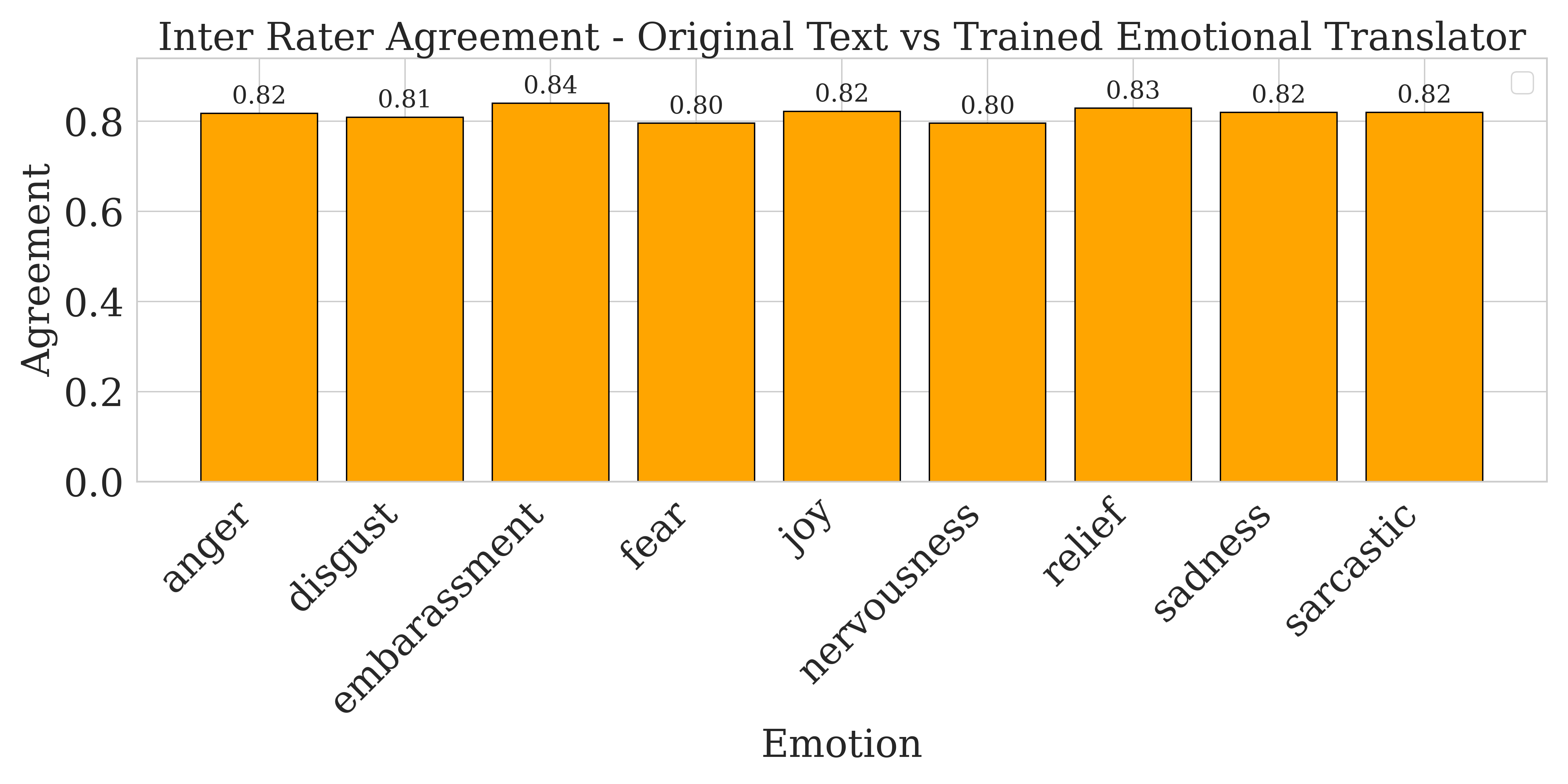}
\caption{Inter-Rater Agreement for comparing emotional fidelity between the original input and the trained intent-translator output. Corresponds to Figure~\ref{fig:more_emotion_original}}
\label{fig:original_vs_trained}
\end{figure}

\begin{figure}[h!]
\centering
\includegraphics[width=\columnwidth]{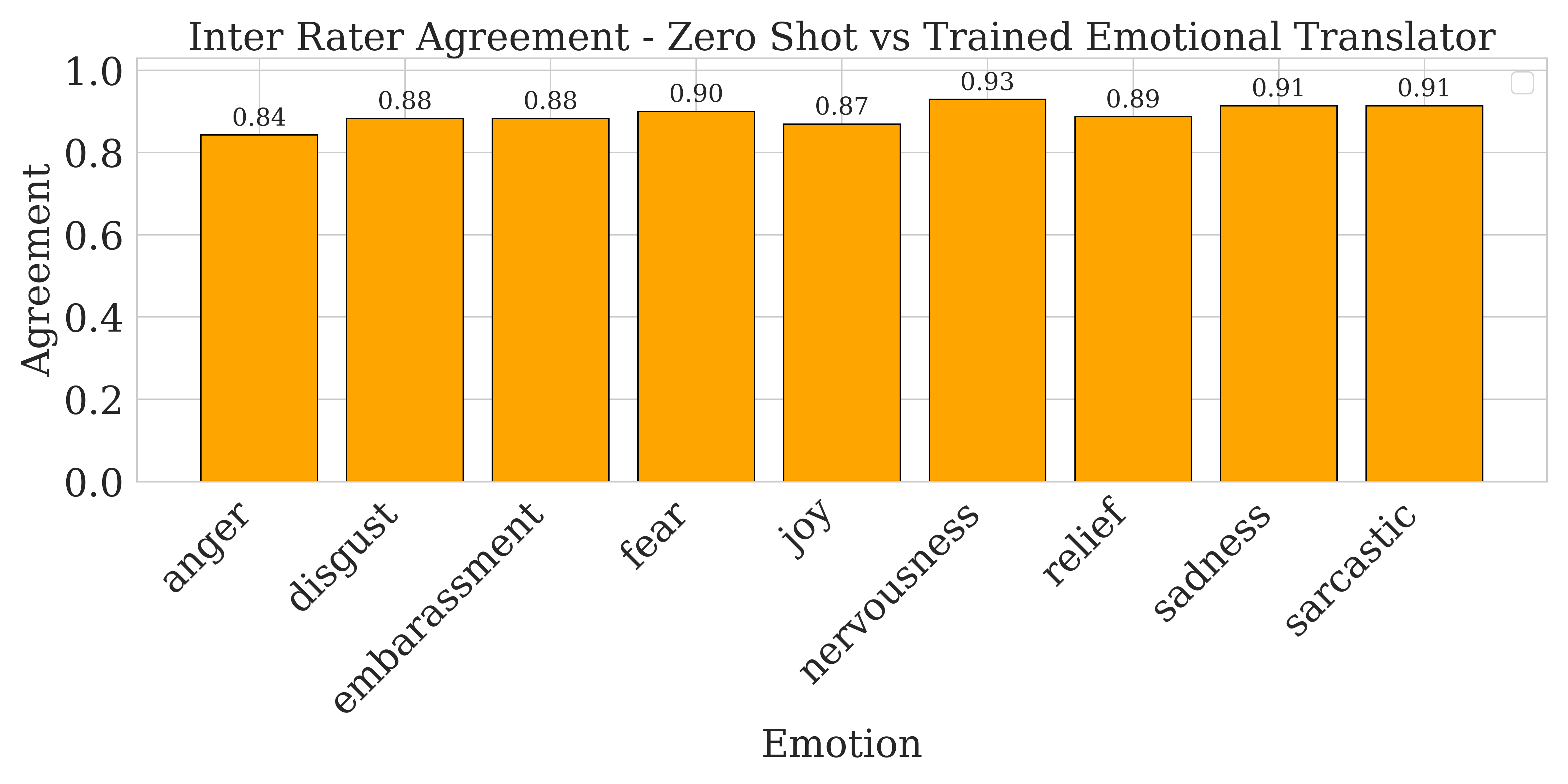}
\caption{Inter-Rater Agreement for comparing emotional and factual quality between the zero-shot LLM and the intent-emotional translator. Corresponds to Figure~\ref{fig:more_emotion_adapter}}
\label{fig:zero_shot_vs_trained}
\end{figure}

\begin{figure}[h!]
\centering
\includegraphics[width=\columnwidth]{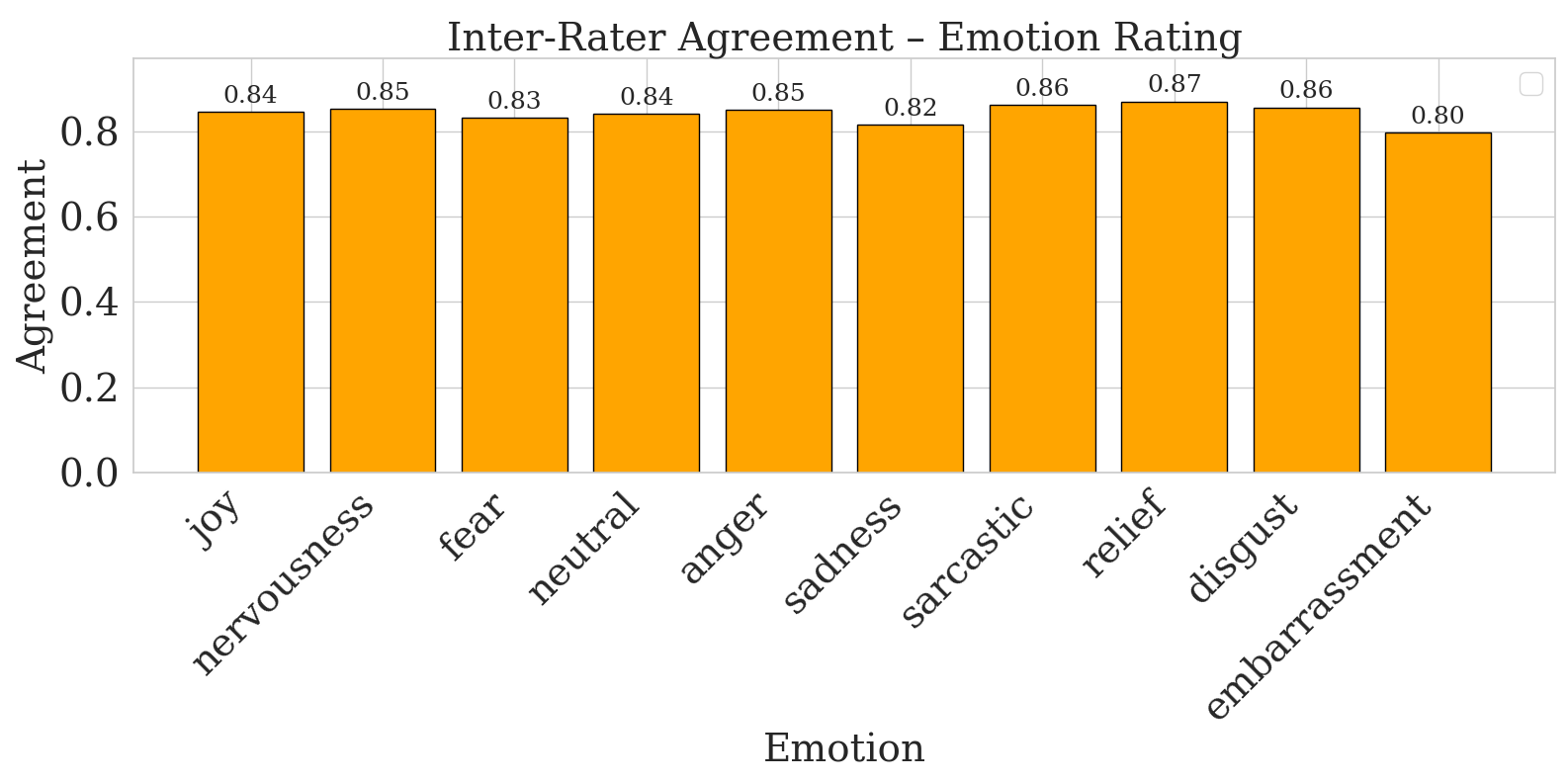}
\caption{Inter-Rater Agreement for evaluating the emotional output of direct-translation results of the intent-translator. Corresponds to Figure~\ref{fig:emotion_translation_direct}}
\label{fig:dt_emotion_rating}
\end{figure}

\begin{figure}[h!]
\centering
\includegraphics[width=\columnwidth]{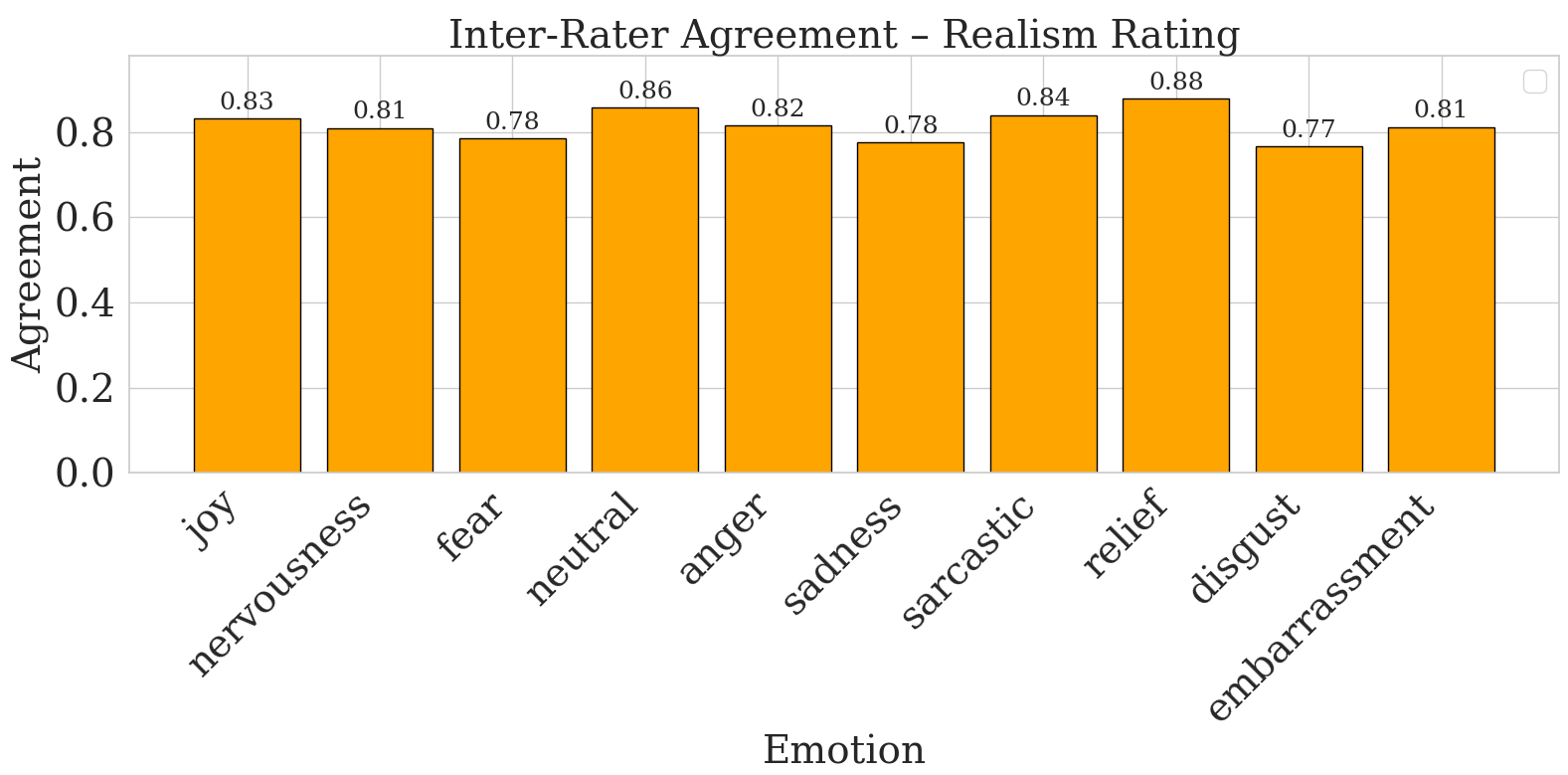}
\caption{Inter-Rater Agreement for evaluating the realism of the output of the direct-translation results of the intent-translator. Corresponds to Figure~\ref{fig:emotion_translation_direct}}
\label{fig:dt_realism_rating}
\end{figure}

\begin{figure}[h!]
\centering
\includegraphics[width=\columnwidth]{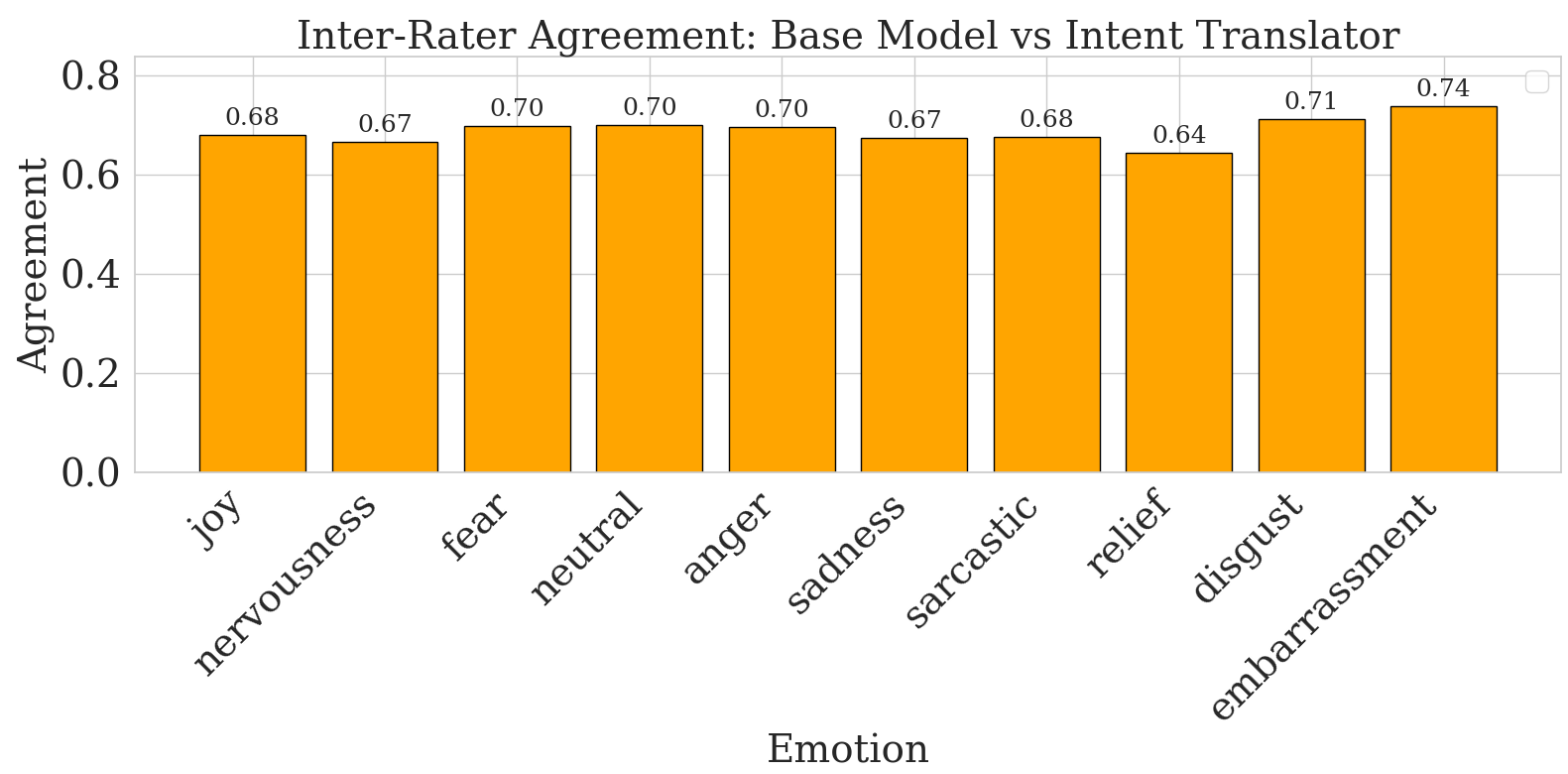}
\caption{Inter-Rater Agreement for model preference between the base model and intent-translator for the direct-translation task. Corresponds to Figure~\ref{fig:model_preference}}
\label{fig:dt_model_comparison}
\end{figure}

\begin{figure}[h!]
\centering
\includegraphics[width=\columnwidth]{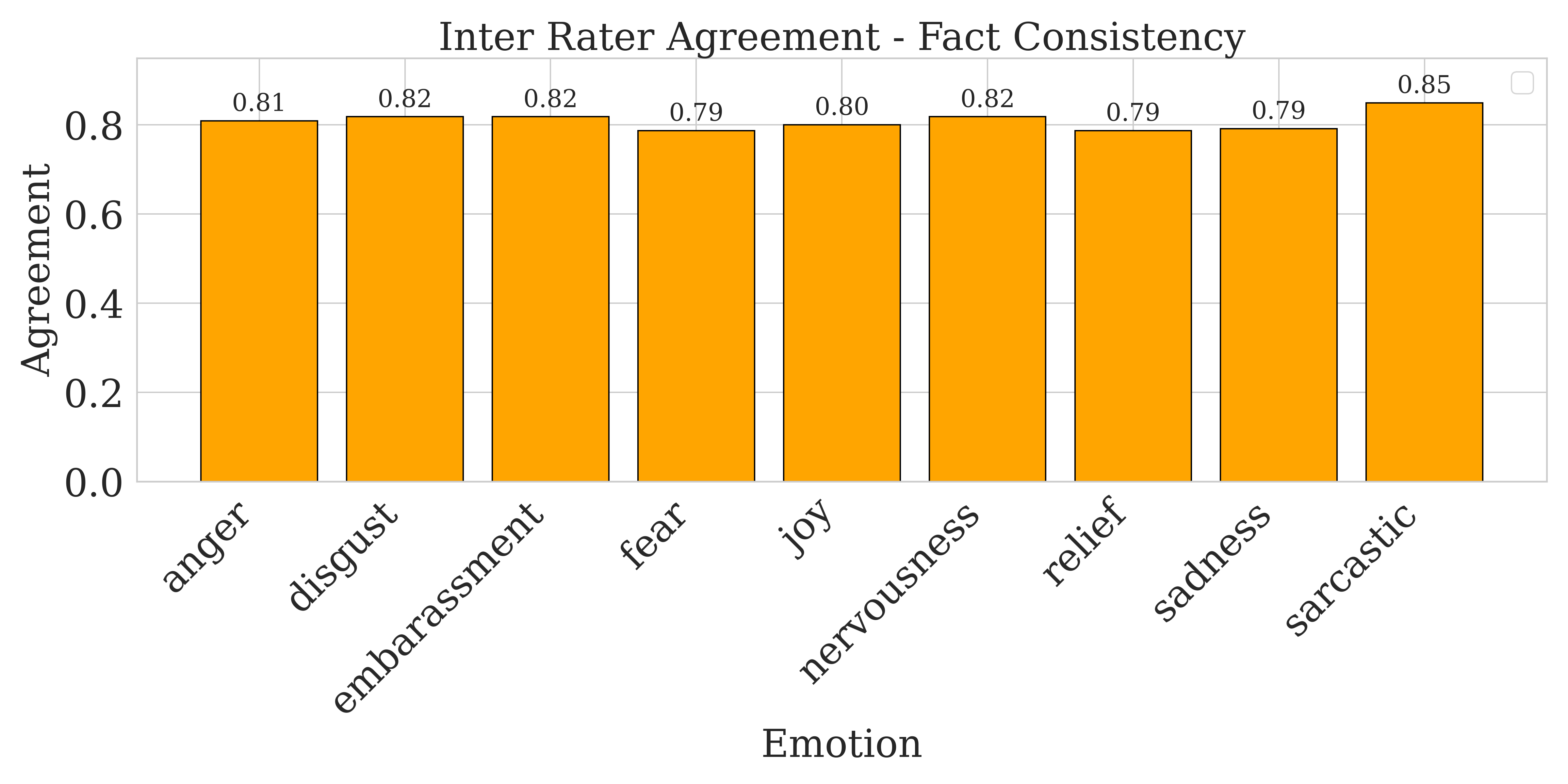}
\caption{Inter-Rater Agreement for evaluating factual consistency in outputs between the zero-shot LLM and the intent-translator text. Corresponds to Figure~\ref{fig:content_reconstruction}}
\label{fig:fact_consistency_eval}
\end{figure}

We report the average inter-rater consistency for each evaluation in Figures~\ref{fig:original_vs_trained}, \ref{fig:zero_shot_vs_trained}, \ref{fig:dt_emotion_rating}, \ref{fig:dt_realism_rating}, \ref{fig:dt_model_comparison}, and \ref{fig:fact_consistency_eval}. Each of these corresponds to a main-text figure:
\begin{itemize}
    \item Figure~\ref{fig:original_vs_trained} aligns with Figure~\ref{fig:more_emotion_original}
    \item Figure~\ref{fig:zero_shot_vs_trained} with Figure~\ref{fig:more_emotion_adapter}
    \item Figures~\ref{fig:dt_emotion_rating} and~\ref{fig:dt_realism_rating} with Figure~\ref{fig:emotion_translation_direct}
    \item Figure~\ref{fig:dt_model_comparison} with Figure~\ref{fig:model_preference}
    \item and Figure~\ref{fig:fact_consistency_eval} with Figure~\ref{fig:content_reconstruction}
\end{itemize}
This provides an intuitive measure of annotator consensus, although it does not account for agreement expected by chance.

\end{document}